\documentclass[twocolumn,english]{IEEEtran}
\usepackage[T1]{fontenc}
\usepackage{color}
\usepackage{babel}
\usepackage{float}
\usepackage{amsthm}
\usepackage{amsmath}
\usepackage{graphicx}
\usepackage[unicode=true,
 bookmarks=true,bookmarksnumbered=true,bookmarksopen=true,bookmarksopenlevel=1,
 breaklinks=false,pdfborder={0 0 0},backref=false,colorlinks=false]
 {hyperref}
\hypersetup{pdftitle={Your Title},
 pdfauthor={Your Name},
 pdfpagelayout=OneColumn, pdfnewwindow=true, pdfstartview=XYZ, plainpages=false}
\usepackage{breakurl}

\makeatletter

\providecommand{\tabularnewline}{\\}
\floatstyle{ruled}
\newfloat{algorithm}{tbp}{loa}
\providecommand{\algorithmname}{Algorithm}
\floatname{algorithm}{\protect\algorithmname}

\theoremstyle{plain}
\newtheorem{thm}{\protect\theoremname}
\theoremstyle{definition}
\newtheorem{example}[thm]{\protect\examplename}
\theoremstyle{plain}
\newtheorem{lem}[thm]{\protect\lemmaname}
\theoremstyle{plain}
\newtheorem{prop}[thm]{\protect\propositionname}
 \usepackage{algolyx}

\ifCLASSOPTIONcompsoc
\usepackage[caption=false,font=normalsize,labelfont=sf,textfont=sf]{subfig}
\else
\usepackage[caption=false,font=footnotesize]{subfig}
\fi

\keycomment{\#<}{>\#}

\newtheorem{assumption}{Assumption}

\makeatother

\providecommand{\examplename}{Example}
\providecommand{\lemmaname}{Lemma}
\providecommand{\propositionname}{Proposition}
\providecommand{\theoremname}{Theorem}

\begin{document}

\title{Sparse Generalized Eigenvalue Problem via Smooth Optimization}

\author{Junxiao Song, Prabhu Babu, and~Daniel P. Palomar,~\IEEEmembership{Fellow,~IEEE}%
\thanks{Junxiao Song, Prabhu Babu, and Daniel P. Palomar are with the Hong
Kong University of Science and Technology (HKUST), Hong Kong. E-mail:
\{jsong, eeprabhubabu, palomar\}@ust.hk.%
}}
\maketitle
\begin{abstract}
In this paper, we consider an $\ell_{0}$-norm penalized formulation
of the generalized eigenvalue problem (GEP), aimed at extracting the
leading sparse generalized eigenvector of a matrix pair. The formulation
involves maximization of a discontinuous nonconcave objective function
over a nonconvex constraint set, and is therefore computationally
intractable. To tackle the problem, we first approximate the $\ell_{0}$-norm
by a continuous surrogate function. Then an algorithm is developed
via iteratively majorizing the surrogate function by a quadratic separable
function, which at each iteration reduces to a regular generalized
eigenvalue problem. A preconditioned steepest ascent algorithm for
finding the leading generalized eigenvector is provided. A systematic
way based on smoothing is proposed to deal with the ``singularity
issue'' that arises when a quadratic function is used to majorize
the nondifferentiable surrogate function. For sparse GEPs with special
structure, algorithms that admit a closed-form solution at every iteration
are derived. Numerical experiments show that the proposed algorithms
match or outperform existing algorithms in terms of computational
complexity and support recovery.\end{abstract}

\begin{IEEEkeywords}
Minorization-maximization, sparse generalized eigenvalue problem,
sparse PCA, smooth optimization.
\end{IEEEkeywords}

\section{Introduction}

\IEEEPARstart{T}{he} generalized eigenvalue problem (GEP) for matrix
pair $(\mathbf{A},\mathbf{B})$ is the problem of finding a pair $(\lambda,\mathbf{x})$
such that 
\begin{equation}
\mathbf{A}\mathbf{x}=\lambda\mathbf{B}\mathbf{x},\label{eq:GEP}
\end{equation}
where $\mathbf{A},$ $\mathbf{B}\in\mathbf{R}^{n\times n},$ $\lambda\in\mathbf{R}$
is called the generalized eigenvalue and $\mathbf{x}\in\mathbf{R}^{n},\mathbf{x}\neq\mathbf{0}$
is the corresponding generalized eigenvector. When $\mathbf{B}$ is
the identity matrix, the problem in \eqref{eq:GEP} reduces to the
simple eigenvalue problem. 

GEP is extremely useful in numerous applications of high dimensional
data analysis and machine learning. Many widely used data analysis
tools, such as principle component analysis (PCA) and canonical correlation
analysis (CCA), are special instances of the generalized eigenvalue
problem \cite{jolliffe1986principal,hotelling1936CCA}. In these applications,
usually $\mathbf{A}\in\mathbf{S}^{n},$ $\mathbf{B}\in\mathbf{S}_{++}^{n}$
(i.e., $\mathbf{A}$ is symmetric and $\mathbf{B}$ is positive definite)
and only a few of the largest generalized eigenvalues are of interest.
In this case, all generalized eigenvalues $\lambda$ and generalized
eigenvectors $\mathbf{x}$ are real and the largest generalized eigenvalue
can be formulated as the following optimization problem
\begin{equation}
\lambda_{{\rm max}}(\mathbf{A},\mathbf{B})=\underset{\mathbf{x}\neq\mathbf{0}}{\mathsf{max}}\,\frac{\mathbf{x}^{T}\mathbf{A}\mathbf{x}}{\mathbf{x}^{T}\mathbf{B}\mathbf{x}},
\end{equation}
or equivalently

\begin{equation}
\lambda_{{\rm max}}(\mathbf{A},\mathbf{B})=\underset{\mathbf{x}}{\mathsf{max}}\left\{ \mathbf{x}^{T}\mathbf{A}\mathbf{x}:\mathbf{x}^{T}\mathbf{B}\mathbf{x}=1\right\} .\label{eq:LGEP}
\end{equation}

Despite the simplicity and popularity of the tools based on GEP, there
is a potential problem: in general the eigenvector is not expected
to have many zero entries, which makes the result difficult to interpret,
especially when dealing with high dimensional data. An ad hoc approach
to fix this problem is to set the entries with absolute values smaller
than a threshold to zero. This thresholding approach is frequently
used in practice, but it is found to be potentially misleading, since
no care is taken on how well the artificially enforced sparsity fits
the original data \cite{cadima1995loading}. Obviously, approaches
that can simultaneously produce accurate and sparse models are more
desirable. 

This has motivated active research in developing methods that enforce
sparsity on eigenvectors, and many approaches have been proposed,
especially for the simple sparse PCA case. For instance, Zou, Hastie,
and Tibshirani \cite{zou2006sparse} first recast the PCA problem
as a ridge regression problem and then imposed $\ell_{1}$-norm penalty
to encourage sparsity. In \cite{d2007direct}, d'Aspremont et al.
proposed a convex relaxation for the sparse PCA problem based on semidefinite
programming (SDP) and Nesterov's smooth minimization technique was
applied to solve the SDP. Shen and Huang \cite{shen2008sparse} exploited
the connection of PCA with singular value decomposition (SVD) of the
data matrix and extracted the sparse principal components (PCs) through
solving a regularized low rank matrix approximation problem. Journée
et al. \cite{Gpower} rewrote the sparse PCA problem in the form of
an optimization problem involving maximization of a convex function
on a compact set and the simple gradient method was then applied.
Although derived differently, the resulting algorithm GPower turns
out to be identical to the rSVD algorithm in \cite{shen2008sparse},
except for the initialization and post-processing phases. Very recently,
Luss and Teboulle \cite{luss2013conditional} introduced an algorithm
framework, called ConGradU, based on the well-known conditional gradient
algorithm, that unifies a variety of seemingly different algorithms,
including the GPower method and the rSVD method. Based on ConGradU,
the authors also proposed a new algorithm for the $\ell_{0}$-constrained
sparse PCA formulation. 

Among the aforementioned algorithms for sparse PCA, rSVD, GPower and
ConGradU are very efficient and require only matrix vector multiplications
at every iteration, thus can be applied to problems of extremely large
size. But these algorithms are not well suited for the case where
$\mathbf{B}$ is not the identity matrix, for example, the sparse
CCA problem, and direct application of these algorithms does not yield
a simple closed-form solution at each iteration any more. To deal
with this problem, \cite{witten2009PMD} suggested that good results
could still be obtained by substituting in the identity matrix for
$\mathbf{B}$ and, in \cite{luss2013conditional}, the authors proposed
to substitute the matrix $\mathbf{B}$ with its diagonal instead.
In \cite{sriperumbudur2007sparse,sriperumbudur2011majorization},
an algorithm was proposed to solve the problem with the general $\mathbf{B}$
(to the best of our knowledge, this is the only one) based on D.C.
(difference of convex functions) programming and minorization-maximization.
The resulting algorithm requires computing a matrix pseudoinverse
and solving a quadratic program (QP) at every iteration when $\mathbf{A}$
is symmetric and positive semidefinite, and in the case where $\mathbf{A}$
is just symmetric it needs to solve a quadratically constrained quadratic
program (QCQP) at each iteration. It is computationally intensive
and not amenable to problems of large size. The same algorithm can
also be applied to the simple sparse PCA problem by simply restricting
$\mathbf{B}$ to be the identity matrix, and in this special case
only one matrix vector multiplication is needed at every iteration
and it is shown to be comparable to the GPower method regarding the
computational complexity.

In this paper, we adopt the MM (majorization-minimization or minorization-maximization)
approach to develop efficient algorithms for the sparse generalized
eigenvalue problem. In fact, all the algorithms that can be unified
by the ConGradU framework can be seen as special cases of the MM method.
Since the ConGradU framework is based on maximizing a convex function
over a compact set via linearizing the convex objective, and the linear
function is just a special minorization function of the convex objective.
Instead of only considering linear minorization function, in this
paper we consider quadratic separable minorization that is related
to the well known iteratively reweighted least squares (IRLS) algorithm
\cite{IRLS1973}. By applying quadratic minorization functions, we
turn the original sparse generalized eigenvalue problem into a sequence
of regular generalized eigenvalue problems and an efficient preconditioned
steepest ascent algorithm for finding the leading generalized eigenvector
is provided. We call the resulting algorithm IRQM (iteratively reweighted
quadratic minorization); it is in spirit similar to IRLS which solves
the $\ell_{1}$-norm minimization problem by solving a sequence of
least squares problems. Algorithms of the IRLS type often suffer from
the infamous ``singularity issue'', i.e., when using quadratic majorization
functions for nondifferentiable functions, the variable may get stuck
at a nondifferentiable point \cite{figueiredo2007majorization}. To
deal with this ``singularity issue'', we propose a systematic way
via smoothing the nondifferentiable surrogate function, which is inspired
by Nesterov's smooth minimization technique for nonsmooth convex optimization
\cite{nesterov2005smooth}, although in our case the surrogate function
is nonconvex. The smoothed problem is shown to be equivalent to a
problem that maximizes a convex objective over a convex constraint
set and the convergence of the IRQM algorithm to a stationary point
of the equivalent problem is proved. For some sparse generalized eigenvalue
problems with special structure, more efficient algorithms are also
derived which admit a closed-form solution at every iteration. 

The remaining sections of the paper are organized as follows. In Section
\ref{sec:Problem-formulation}, the problem formulation of the sparse
generalized eigenvalue problem is presented and the surrogate functions
that will be used to approximate $\ell_{0}$-norm are discussed. In
Section \ref{sec:Sparse-Generalized-Eigenvalue}, we first give a
brief review of the MM framework and then algorithms based on the
MM framework are derived for the sparse generalized eigenvalue problems
in general and with special structure. A systematic way to deal with
the ``singularity issue'' arising when using quadratic minorization
functions is also proposed. In Section \ref{sec:Convergence-analysis},
the convergence of the proposed MM algorithms is analyzed. Section
\ref{sec:Numerical-Experiments} presents numerical experiments and
the conclusions are given in Section \ref{sec:Conclusions}.

\emph{Notation}: $\mathbf{R}$ and $\mathbf{C}$ denote the real field
and the complex field, respectively. $\mathrm{Re}(\cdot)$ and $\mathrm{Im}(\cdot)$
denote the real and imaginary part, respectively. $\mathbf{R}^{n}$
$(\mathbf{R}_{+}^{n},\mathbf{R}_{++}^{n})$ denotes the set of (nonnegative,
strictly positive) real vectors of size $n.$ $\mathbf{S}^{n}$ $(\mathbf{S}_{+}^{n},\mathbf{S}_{++}^{n})$
denotes the set of symmetric (positive semidefinite, positive definite)
$n\times n$ matrices defined over $\mathbf{R}$. Boldface upper case
letters denote matrices, boldface lower case letters denote column
vectors, and italics denote scalars. The superscripts $(\cdot)^{T}$
and $(\cdot)^{H}$ denote transpose and conjugate transpose, respectively.
$X_{i,j}$ denotes the (\emph{i}-th, \emph{j}-th) element of matrix
$\mathbf{X}$ and $x_{i}$ denotes the \emph{i}-th element of vector
$\mathbf{x}$. $\mathbf{X}_{i,:}$ denotes the \emph{i}-th row of
matrix $\mathbf{X}$, $\mathbf{X}_{:,j}$ denotes the \emph{j}-th
column of matrix $\mathbf{X}$. ${\sf diag}(\mathbf{X})$ is a column
vector consisting of all the diagonal elements of $\mathbf{X}$. ${\rm Diag}(\mathbf{x})$
is a diagonal matrix formed with $\mathbf{x}$ as its principal diagonal.
Given a vector $\mathbf{x}\in\mathbf{R}^{n},$ $\left|\mathbf{x}\right|$
denotes the vector with $i$th entry $\left|x_{i}\right|,$ $\left\Vert \mathbf{x}\right\Vert _{0}$
denotes the number of non-zero elements of $\mathbf{x}$, $\left\Vert \mathbf{x}\right\Vert _{p}:=\left(\sum_{i=1}^{n}\left|x_{i}\right|^{p}\right)^{1/p},0<p<\infty.$
$\mathbf{I}_{n}$ denotes an $n\times n$ identity matrix. ${\rm sgn}(x)$
denotes the sign function, which takes $-1,0,1$ if $x<0,x=0,x>0$,
respectively.

\section{Problem formulation \label{sec:Problem-formulation}}

Given a symmetric matrix $\mathbf{A}\in\mathbf{S}^{n}$ and a symmetric
positive definite matrix $\mathbf{B}\in\mathbf{S}_{++}^{n},$ the
main problem of interest is the following $\text{\ensuremath{\ell}}_{0}$-norm
regularized generalized eigenvalue problem
\begin{equation}
\begin{array}{ll}
\underset{\mathbf{x}}{\mathsf{maximize}} & \mathbf{x}^{T}\mathbf{A}\mathbf{x}-\rho\left\Vert \mathbf{x}\right\Vert _{0}\\
\mathsf{subject\; to} & \mathbf{x}^{T}\mathbf{B}\mathbf{x}=1,
\end{array}\label{eq:SGEP-L0}
\end{equation}
where $\rho>0$ is the regularization parameter. This formulation
is general enough and includes some sparse PCA and sparse CCA formulations
in the literature as special cases. \textcolor{red}{}

\begin{figure}[htbp]
\centering{}\includegraphics[width=0.7\columnwidth]{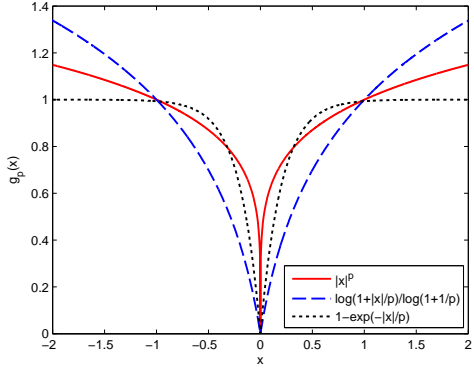}
\protect\caption{\label{fig:surrogate_functions}Three surrogate functions $g_{p}(x)$
that are used to approximate $\mathrm{sgn}(\left|x\right|)$, $p=0.2$.}
\end{figure}

The problem \eqref{eq:SGEP-L0} involves the maximization of a non-concave
discontinuous objective over a nonconvex set, thus really hard to
deal with directly. The intractability of the problem is not only
due to the nonconvexity, but also due to the discontinuity of the
cardinality function in the objective. A natural approach to deal
with the discontinuity of the $\ell_{0}$-norm is to approximate it
by some continuous function. It is easy to see that the $\ell_{0}$-norm
can be written as 
\[
\left\Vert \mathbf{x}\right\Vert _{0}=\sum_{i=1}^{n}{\rm sgn}(\left|x_{i}\right|).
\]
Thus, to approximate $\left\Vert \mathbf{x}\right\Vert _{0},$ we
may just replace the problematic ${\rm sgn}(\left|x_{i}\right|)$
by some nicer surrogate function $g_{p}(x_{i}),$ where $p>0$ is
a parameter that controls the approximation. In this paper, we will
consider the class of continuous even functions defined on $\mathbf{R}$,
which are differentiable everywhere except at $0$ and concave and
monotone increasing on $[0,+\infty)$ and $g_{p}(0)=0$. In particular,
we will consider the following three surrogate functions:
\begin{enumerate}
\item $g_{p}(x)=\left|x\right|^{p},$ $0<p\leq1$
\item $g_{p}(x)=\log(1+\left|x\right|/p)$/$\log(1+1/p)$, $p>0$
\item $g_{p}(x)=1-e^{-\left|x\right|/p},$ $p>0.$
\end{enumerate}
The first is the $p$-norm-like measure (with $p\leq1$) used in \cite{FOCUSS1997,chartrand2008IRLS},
which is shown to perform well in promoting sparse solutions for compressed
sensing problems. The second is the penalty function used in \cite{sriperumbudur2011majorization}
for sparse generalized eigenvalue problem and when used to replace
the $\ell_{1}$-norm in basis pursuit, it leads to the well known
iteratively reweighted $\ell_{1}$-norm minimization algorithm \cite{candes2008enhancing}.
The last surrogate function is used in \cite{mangasarian1996machine}
for feature selection problems, which is different from the first
two surrogate functions in the sense that it has the additional property
of being a lower bound of the function $\mathrm{sgn}(\left|x\right|)$.
To provide an intuitive idea about how these surrogate functions look
like, they are plotted in Fig. \ref{fig:surrogate_functions} for
fixed $p=0.2.$

By approximating $\left\Vert \mathbf{x}\right\Vert _{0}$ with $\sum_{i=1}^{n}g_{p}(x_{i})$,
the original problem \eqref{eq:SGEP-L0} is approximated by the following
problem
\begin{equation}
\begin{array}{ll}
\underset{\mathbf{x}}{\mathsf{maximize}} & \mathbf{x}^{T}\mathbf{A}\mathbf{x}-\rho\sum_{i=1}^{n}g_{p}(x_{i})\\
\mathsf{subject\; to} & \mathbf{x}^{T}\mathbf{B}\mathbf{x}=1.
\end{array}\label{eq:SGEP-Approx}
\end{equation}
With the approximation, the problem \eqref{eq:SGEP-Approx} is still
a nonconvex nondifferentiable optimization problem, but it is a continuous
problem now in contrast to the original problem \eqref{eq:SGEP-L0}.
In the following section, we will concentrate on the approximate problem
\eqref{eq:SGEP-Approx} and develop fast algorithms to solve it based
on the MM (majorization-minimization or minorization-maximization)
scheme. 

Note that for simplicity of exposition, we focus on real-valued matrices
throughout the paper. However, the techniques developed in this paper
can be adapted for complex-valued matrix pair $(\mathbf{A},\mathbf{B})$,
with $\mathbf{A}$ being an $n\times n$ Hermitian matrix and $\mathbf{B}$
being an $n\times n$ Hermitian positive definite matrix. One approach
is to transform the problem to a real-valued one by defining 
\[
\tilde{\mathbf{x}}=[{\rm Re}(\mathbf{x})^{T},{\rm Im}(\mathbf{x})^{T}]^{T}
\]
\[
\tilde{\mathbf{A}}=\left[\begin{array}{cc}
{\rm Re}(\mathbf{A}) & -{\rm Im}(\mathbf{A})\\
{\rm Im}(\mathbf{A}) & {\rm Re}(\mathbf{A})
\end{array}\right],\,\tilde{\mathbf{B}}=\left[\begin{array}{cc}
{\rm Re}(\mathbf{B}) & -{\rm Im}(\mathbf{B})\\
{\rm Im}(\mathbf{B}) & {\rm Re}(\mathbf{B})
\end{array}\right].
\]
In this approach, the cardinality is considered for the real and imaginary
part of the vector $\mathbf{x}$ separately. A more natural approach
is to consider directly the complex-valued version of the $\text{\ensuremath{\ell}}_{0}$-norm
regularized generalized eigenvalue problem 
\[
\begin{array}{ll}
\underset{\mathbf{x}\in\mathbf{C}^{n}}{\mathsf{maximize}} & \mathbf{x}^{H}\mathbf{A}\mathbf{x}-\rho\left\Vert \mathbf{x}\right\Vert _{0}\\
\mathsf{subject\; to} & \mathbf{x}^{H}\mathbf{B}\mathbf{x}=1,
\end{array}
\]
where the $\text{\ensuremath{\ell}}_{0}$-norm can still be written
as $\left\Vert \mathbf{x}\right\Vert _{0}=\sum_{i=1}^{n}{\rm sgn}(\left|x_{i}\right|)$,
but with $\left|x_{i}\right|$ being the modulus of $x_{i}$ now.
Notice that the three surrogate functions $g_{p}(x)$ used to approximate
${\rm sgn}(\left|x\right|)$ are all functions of $\left|x\right|$,
by taking $\left|x\right|$ as the modulus of a complex number, the
surrogate functions are directly applicable to the complex case. The
quadratic minorization function that will be described in Section
\ref{sec:Sparse-Generalized-Eigenvalue} can also be constructed similarly
in the complex case and at each iteration of the resulting algorithm
we still need to solve a regular generalized eigenvalue problem but
with complex-valued matrices.

\section{Sparse Generalized Eigenvalue Problem via MM Scheme\label{sec:Sparse-Generalized-Eigenvalue} }

\subsection{The MM method\label{sub:MM-method}}

The MM method refers to the majorization-minimization method or the
minorization-maximization method, which is a generalization of the
well known expectation maximization (EM) algorithm. It is an approach
to solve optimization problems that are too difficult to solve directly.
The principle behind the MM method is to transform a difficult problem
into a series of simple problems. Interested readers may refer to
\cite{hunter2004MMtutorial} and references therein for more details.

Suppose we want to minimize $f(\mathbf{x})$ over $\mathcal{X}\in\mathbf{R}^{n}$.
Instead of minimizing the cost function $f(\mathbf{x})$ directly,
the MM approach optimizes a sequence of approximate objective functions
that majorize $f(\mathbf{x})$. More specifically, starting from a
feasible point $\mathbf{x}^{(0)},$ the algorithm produces a sequence
$\{\mathbf{x}^{(k)}\}$ according to the following update rule 
\begin{equation}
\mathbf{x}^{(k+1)}\in\underset{\mathbf{x}\in\mathcal{X}}{\arg\min}\,\, u(\mathbf{x},\mathbf{x}^{(k)}),\label{eq:major_update}
\end{equation}
where $\mathbf{x}^{(k)}$ is the point generated by the algorithm
at iteration $k,$ and $u(\mathbf{x},\mathbf{x}^{(k)})$ is the majorization
function of $f(\mathbf{x})$ at $\mathbf{x}^{(k)}$. Formally, the
function $u(\mathbf{x},\mathbf{x}^{(k)})$ is said to majorize the
function $f(\mathbf{x})$ at the point $\mathbf{x}^{(k)}$ provided
\begin{eqnarray}
u(\mathbf{x},\mathbf{x}^{(k)}) & \geq & f(\mathbf{x}),\quad\forall\mathbf{x}\in\mathcal{X},\label{eq:major1}\\
u(\mathbf{x}^{(k)},\mathbf{x}^{(k)}) & = & f(\mathbf{x}^{(k)}).\label{eq:major2}
\end{eqnarray}
In other words, function $u(\mathbf{x},\mathbf{x}^{(k)})$ is a global
upper bound for $f(\mathbf{x})$ and coincides with $f(\mathbf{x})$
at $\mathbf{x}^{(k)}$. 

It is easy to show that with this scheme, the objective value is decreased
 monotonically at every iteration, i.e., 
\begin{equation}
f(\mathbf{x}^{(k+1)})\leq u(\mathbf{x}^{(k+1)},\mathbf{x}^{(k)})\leq u(\mathbf{x}^{(k)},\mathbf{x}^{(k)})=f(\mathbf{x}^{(k)}).\label{eq:descent-property}
\end{equation}
The first inequality and the third equality follow from the the properties
of the majorization function, namely \eqref{eq:major1} and \eqref{eq:major2}
respectively and the second inequality follows from \eqref{eq:major_update}.

Note that with straightforward changes, similar scheme can be applied
to maximization. To maximize a function $f(\mathbf{x})$, we need
to minorize it by a surrogate function $u(\mathbf{x},\mathbf{x}^{(k)})$
and maximize $u(\mathbf{x},\mathbf{x}^{(k)})$ to produce the next
iterate $\mathbf{x}^{(k+1)}.$ A function $u(\mathbf{x},\mathbf{x}^{(k)})$
is said to minorize the function $f(\mathbf{x})$ at the point $\mathbf{x}^{(k)}$
if $-u(\mathbf{x},\mathbf{x}^{(k)})$ majorizes $-f(\mathbf{x})$
at $\mathbf{x}^{(k)}$. This scheme refers to minorization-maximization
and similarly it is easy to shown that with this scheme the objective
value is increased at each iteration.

\subsection{Quadratic Minorization Function\label{sub:Quadratic-Minorization-Function}}

\begin{figure}[htbp]
\centering{}\includegraphics[width=0.68\columnwidth]{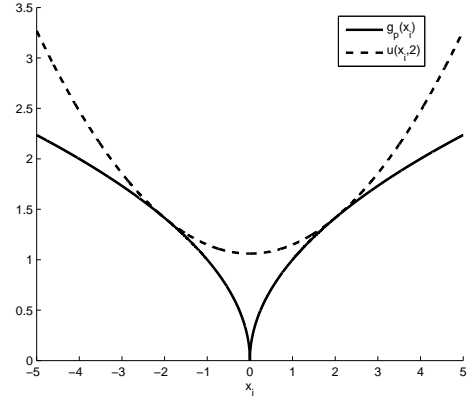}
\protect\caption{\label{fig:majorizer}The function $g_{p}(x_{i})=\left|x_{i}\right|^{p}$
with $p=0.5$ and its quadratic majorization function at $x_{i}^{(k)}=2.$}
\end{figure}

Having briefly introduced the general MM framework, let us return
to the approximate sparse generalized eigenvalue problem (SGEP) in
\eqref{eq:SGEP-Approx}. To apply the MM scheme, the key step is to
find an appropriate minorization function for the objective of \eqref{eq:SGEP-Approx}
at each iteration such that the resulting problem is easy to solve.
To construct such a minorization function, we keep the quadratic term
$\mathbf{x}^{T}\mathbf{A}\mathbf{x}$ and only minorize the penalty
term $-\rho\sum_{i=1}^{n}g_{p}(x_{i})$ (i.e., majorize $\rho\sum_{i=1}^{n}g_{p}(x_{i})$).
More specifically, at iteration $k$ we majorize each of the surrogate
functions $g_{p}(x_{i}),i=1,\ldots,n$ at $x_{i}^{(k)}$ with a quadratic
function $w_{i}^{(k)}x_{i}^{2}+c_{i}^{(k)}$, where the coefficients
$w_{i}^{(k)}$ and $c_{i}^{(k)}$ are determined by the following
two conditions (for $x_{i}^{(k)}\neq0$):
\begin{eqnarray}
g_{p}(x_{i}^{(k)}) & = & w_{i}^{(k)}\left(x_{i}^{(k)}\right)^{2}+c_{i}^{(k)},\label{eq:major-cond1-1}\\
g_{p}^{\prime}(x_{i}^{(k)}) & = & 2w_{i}^{(k)}x_{i}^{(k)},\label{eq:major-cond2-1}
\end{eqnarray}
i.e., the quadratic function coincides with the surrogate function
$g_{p}(x_{i})$ at $x_{i}^{(k)}$ and is also tangent to $g_{p}(x_{i})$
at $x_{i}^{(k)}$. Due to the fact that the surrogate functions of
interest are differentiable and concave for $x_{i}>0$ (also for $x_{i}<0$),
the second condition implies that the quadratic function is a global
upper bound of the surrogate function $g_{p}(x_{i})$. Then the objective
of \eqref{eq:SGEP-Approx}, i.e., $\mathbf{x}^{T}\mathbf{A}\mathbf{x}-\rho\sum_{i=1}^{n}g_{p}(x_{i}),$
is minorized by the quadratic function $\mathbf{x}^{T}\mathbf{A}\mathbf{x}-\rho\sum_{i=1}^{n}\left(w_{i}^{(k)}x_{i}^{2}+c_{i}^{(k)}\right),$
which can be written more compactly as 
\begin{equation}
\mathbf{x}^{T}\left(\mathbf{A}-\rho{\rm Diag}(\mathbf{w}^{(k)})\right)\mathbf{x}-\rho\sum_{i=1}^{n}c_{i}^{(k)},\label{eq:obj_majorizer}
\end{equation}
where $\mathbf{w}^{(k)}=[w_{1}^{(k)},\ldots,w_{n}^{(k)}]^{T}.$
\begin{example}
To compute the quadratic function $w_{i}^{(k)}x_{i}^{2}+c_{i}^{(k)}$
that majorizes the surrogate function $g_{p}(x_{i})=\left|x_{i}\right|^{p},0<p\leq1$
at $x_{i}^{(k)}\neq0$, we have the following two equations corresponding
to \eqref{eq:major-cond1-1} and \eqref{eq:major-cond2-1} respectively:
\begin{eqnarray}
\left|x_{i}^{(k)}\right|^{p} & = & w_{i}^{(k)}\left(x_{i}^{(k)}\right)^{2}+c_{i}^{(k)},\label{eq:major-cond1-example}\\
{\rm sgn}(x_{i}^{(k)})p\left|x_{i}^{(k)}\right|^{p-1} & = & 2w_{i}^{(k)}x_{i}^{(k)}.\label{eq:major-cond2-example}
\end{eqnarray}
By solving \eqref{eq:major-cond1-example} and \eqref{eq:major-cond2-example},
we can get the quadratic majorization function
\begin{equation}
u(x_{i},x_{i}^{(k)})=\frac{p}{2}\left|x_{i}^{(k)}\right|^{p-2}x_{i}^{2}+(1-\frac{p}{2})\left|x_{i}^{(k)}\right|^{p},\label{eq:quadratic_major_function}
\end{equation}
which is illustrated in Fig. \ref{fig:majorizer} with $p=0.5$ and
$x_{i}^{(k)}=2.$
\end{example}

In fact, the idea of majorizing some penalty functions by quadratic
separable functions is well known in robust regression \cite{IRLSholland1977robust}.
It was first proposed to solve the absolute deviations curve fitting
problem (i.e., regression with $\ell_{1}$-norm cost function) via
iteratively solving a series of weighted least squares problems and
the resulting algorithm is known as the iteratively reweighted least
squares (IRLS) algorithm \cite{IRLS1973}. Later the idea was applied
in signal processing for sparse signal reconstruction in \cite{FOCUSS1997,rao1999FOCUSS}.
Recently, the IRLS approach has also been applied in Compressed Sensing
\cite{chartrand2008IRLS}. Algorithms based on this idea often have
the infamous singularity issue \cite{figueiredo2007majorization},
that is, the quadratic majorization function $w_{i}^{(k)}x_{i}^{2}+c_{i}^{(k)}$
is not defined at $x_{i}^{(k)}=0,$ due to the nondifferentiability
of $g_{p}(x_{i})$ at $x_{i}=0.$ For example, considering the quadratic
majorization function of $g_{p}(x_{i})=\left|x_{i}\right|^{p},0<p\leq1$
in \eqref{eq:quadratic_major_function}, it is easy to see that the
coefficient $\frac{p}{2}\left|x_{i}^{(k)}\right|^{p-2}$ is not defined
at $x_{i}^{(k)}=0$. To tackle this problem, the authors of \cite{figueiredo2007majorization}
proposed to define the majorization function at the particular point
$x_{i}^{(k)}=0$ as 
\[
u(x_{i},0)=\begin{cases}
+\infty, & x_{i}\neq0\\
0, & x_{i}=0,
\end{cases}
\]
which implies that once $x_{i}^{(k)}=0$, the next iteration will
also be zero, i.e., $x_{i}^{(k+1)}=0$. This may impact the convergence
of the algorithm to a minimizer of the objective, if the corresponding
element of the minimizer is in fact not zero. Another common approach
for dealing with this singularity issue is to incorporate a small
$\epsilon>0$, for example, for the quadratic majorization function
of $g_{p}(x_{i})=\left|x_{i}\right|^{p},\,0<p\leq1$ in \eqref{eq:quadratic_major_function},
replace the coefficient $\frac{p}{2}\left|x_{i}^{(k)}\right|^{p-2}$
by 
\[
w_{i}^{(k)}=\frac{p}{2}\left(\left(x_{i}^{(k)}\right)^{2}+\epsilon\right)^{\frac{p-2}{2}},
\]
which is the so called damping approach used in \cite{chartrand2008IRLS}.
A potential problem of this approach is that although $\epsilon$
is small, we have no idea how it will affect the convergence of the
algorithm, since the corresponding quadratic function is no longer
a majorization function of the surrogate function.

\subsection{Smooth Approximations of Non-differentiable Surrogate Functions\label{sub:Smooth-Approximations}}

To tackle the singularity issue arised during the construction of
the quadratic minorization function in \eqref{eq:obj_majorizer},
in this subsection we propose to incorporate a small $\epsilon>0$
in a more systematic way. 

The idea is to approximate the non-differentiable surrogate function
$g_{p}(x)$ with a differentiable function of the following form 
\[
g_{p}^{\epsilon}(x)=\begin{cases}
ax^{2}, & \left|x\right|\leq\epsilon\\
g_{p}(x)-b, & \left|x\right|>\epsilon,
\end{cases}
\]
which aims at smoothening the non-differentiable surrogate function
around zero by a quadratic function. To make the function $g_{p}^{\epsilon}(x)$
continuous and differentiable at $x=\pm\epsilon,$ the following two
conditions are needed $a\epsilon^{2}=g_{p}(\epsilon)-b,$ $2a\epsilon=g_{p}^{\prime}(\epsilon)$,
which lead to $a=\frac{g_{p}^{\prime}(\epsilon)}{2\epsilon}$, $b=g_{p}(\epsilon)-\frac{g_{p}^{\prime}(\epsilon)}{2}\epsilon$.
Thus, the smooth approximation of the surrogate function that will
be employed is 
\begin{equation}
g_{p}^{\epsilon}(x)=\begin{cases}
\frac{g_{p}^{\prime}(\epsilon)}{2\epsilon}x^{2}, & \left|x\right|\leq\epsilon\\
g_{p}(x)-g_{p}(\epsilon)+\frac{g_{p}^{\prime}(\epsilon)}{2}\epsilon, & \left|x\right|>\epsilon,
\end{cases}\label{eq:smooth-surrogate}
\end{equation}
where $\epsilon>0$ is a constant parameter.
\begin{example}
The smooth approximation of the function $g_{p}(x)=\left|x\right|^{p},\,0<p\leq1$
is
\begin{equation}
g_{p}^{\epsilon}(x)=\begin{cases}
\frac{p}{2}\epsilon^{p-2}x^{2}, & \left|x\right|\leq\epsilon\\
\left|x\right|^{p}-(1-\frac{p}{2})\epsilon^{p}, & \left|x\right|>\epsilon.
\end{cases}\label{eq:Lp_norm_smooth}
\end{equation}
The case with $p=0.5$ and $\epsilon=0.05$ is illustrated in Fig.
\ref{fig:smoother}. When $p=1$, the smooth approximation \eqref{eq:Lp_norm_smooth}
is the well known Huber penalty function and the application of Huber
penalty as smoothed absolute value has been used in \cite{becker2011NESTA}
to derive fast algorithms for sparse recovery.

\begin{figure}[htbp]
\centering{}\includegraphics[width=0.7\columnwidth]{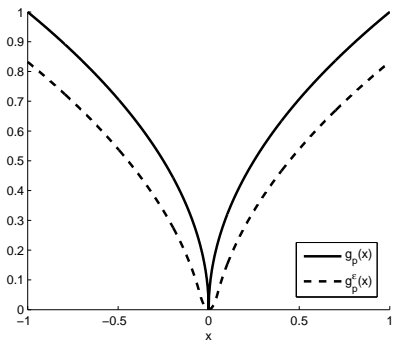}
\protect\caption{\label{fig:smoother}The function $g_{p}(x)=\left|x\right|^{p}$ with
$p=0.5$ and its smooth approximation $g_{p}^{\epsilon}(x)$ with
$\epsilon=0.05$. }
\end{figure}

\end{example}
With this smooth approximation, the problem \eqref{eq:SGEP-Approx}
becomes the following smoothed one: 
\begin{equation}
\begin{array}{ll}
\underset{\mathbf{x}}{\mathsf{maximize}} & \mathbf{x}^{T}\mathbf{A}\mathbf{x}-\rho\sum_{i=1}^{n}g_{p}^{\epsilon}(x_{i})\\
\mathsf{subject\; to} & \mathbf{x}^{T}\mathbf{B}\mathbf{x}=1,
\end{array}\label{eq:SGEP-smooth-Approx}
\end{equation}
where $g_{p}^{\epsilon}(\cdot)$ is the function given by \eqref{eq:smooth-surrogate}.
We now majorize the smoothed surrogate functions $g_{p}^{\epsilon}(x_{i})$
with quadratic functions and the coefficients of the quadratic majorization
functions are summarized in Table \ref{tab:Smooth-surrogate-functions}
(we have omitted the coefficient $c_{i}^{(k)}$ since it is irrelevant
for the MM algorithm). Notice that the quadratic functions $w_{i}^{(k)}x_{i}^{2}+c_{i}^{(k)}$
are now well defined. Thus, the smooth approximation we have constructed
can be viewed as a systematic way to incorporate a small $\epsilon>0$
to deal with the singularity issue of IRLS type algorithms.

\begin{table*}[htbp]
\centering \protect\caption{\label{tab:Smooth-surrogate-functions}Smooth approximation $g_{p}^{\epsilon}(x_{i})$
of the surrogate functions $g_{p}(x_{i})$ and the quadratic majorization
functions $u(x_{i},x_{i}^{(k)})=w_{i}^{(k)}x_{i}^{2}+c_{i}^{(k)}$
at $x_{i}^{(k)}.$}

\centering{}%
\begin{tabular}{|c|c|c|}
\hline 
Surrogate function $g_{p}(x_{i})$ & Smooth approximation $g_{p}^{\epsilon}(x_{i})$  & $w_{i}^{(k)}$\tabularnewline
\hline 
$\left|x_{i}\right|^{p},\,0<p\leq1$ & $\begin{cases}
\frac{p}{2}\epsilon^{p-2}x_{i}^{2}, & \left|x_{i}\right|\leq\epsilon\\
\left|x_{i}\right|^{p}-(1-\frac{p}{2})\epsilon^{p}, & \left|x_{i}\right|>\epsilon
\end{cases}$ & $\begin{cases}
\frac{p}{2}\epsilon^{p-2}, & \big|x_{i}^{(k)}\big|\leq\epsilon\\
\frac{p}{2}\left|x_{i}^{(k)}\right|^{p-2}, & \big|x_{i}^{(k)}\big|>\epsilon
\end{cases}$\tabularnewline
\hline 
$\log(1+\left|x_{i}\right|/p)/\log(1+1/p),\, p>0$ & $\begin{cases}
\frac{x_{i}^{2}}{2\epsilon(p+\epsilon)\log(1+1/p)}, & \left|x_{i}\right|\leq\epsilon\\
\frac{\log(1+\left|x_{i}\right|/p)-\log(1+\epsilon/p)+\frac{\epsilon}{2(p+\epsilon)}}{\log(1+1/p)}, & \left|x_{i}\right|>\epsilon
\end{cases}$ & $\begin{cases}
\frac{1}{2\epsilon(p+\epsilon)\log(1+1/p)}, & \big|x_{i}^{(k)}\big|\leq\epsilon\\
\frac{1}{2\log(1+1/p)\left|x_{i}^{(k)}\right|(\left|x_{i}^{(k)}\right|+p)}, & \big|x_{i}^{(k)}\big|>\epsilon
\end{cases}$\tabularnewline
\hline 
$1-e^{-\left|x_{i}\right|/p}$, $p>0$ & $\begin{cases}
\frac{e^{-\epsilon/p}}{2p\epsilon}x_{i}^{2}, & \left|x_{i}\right|\leq\epsilon\\
-e^{-\left|x_{i}\right|/p}+\left(1+\frac{\epsilon}{2p}\right)e^{-\epsilon/p}, & \left|x_{i}\right|>\epsilon
\end{cases}$ & $\begin{cases}
\frac{e^{-\epsilon/p}}{2p\epsilon}, & \big|x_{i}^{(k)}\big|\leq\epsilon\\
\frac{e^{-\big|x_{i}^{(k)}\big|/p}}{2p\big|x_{i}^{(k)}\big|}, & \big|x_{i}^{(k)}\big|>\epsilon
\end{cases}$\tabularnewline
\hline 
\end{tabular}
\end{table*}

Although there is no singularity issue when applying the quadratic
minorization to the smoothed problem \eqref{eq:SGEP-smooth-Approx},
a natural question is if we solve the smoothed problem, what can we
say about the solution with regard to the original problem \eqref{eq:SGEP-Approx}.
In the following, we present some results which aim at answering the
question.
\begin{lem}
\label{lem:Low-up_bound}Let $g_{p}(\cdot)$ be a continuous even
function defined on $\mathbf{R}$, differentiable everywhere except
at zero, concave and monotone increasing on $[0,+\infty)$ with $g_{p}(0)=0.$
Then the smooth approximation $g_{p}^{\epsilon}(x)$ defined by \eqref{eq:smooth-surrogate}
is a global lower bound of $g_{p}(x)$ and $g_{p}^{\epsilon}(x)+g_{p}(\epsilon)-\frac{g_{p}^{\prime}(\epsilon)}{2}\epsilon$
is a global upper bound of $g_{p}(x)$.\end{lem}
\begin{IEEEproof}
The lemma is quite intuitive and the proof is omitted for lack of
space.\end{IEEEproof}
\begin{prop}
\label{prop:suboptimal}Let $f(\mathbf{x})$ be the objective of the
problem \eqref{eq:SGEP-Approx}, with $g_{p}(\cdot)$ as in Lemma
\ref{lem:Low-up_bound} and $f_{\epsilon}(\mathbf{x})$ be the objective
of the smoothed problem \eqref{eq:SGEP-smooth-Approx} with $g_{p}^{\epsilon}(\cdot)$
as in \eqref{eq:smooth-surrogate}. Let $\mathbf{x}^{\star}$ be the
optimal solution of the problem \eqref{eq:SGEP-Approx} and $\mathbf{x}_{\epsilon}^{\star}$
be the optimal solution of the smoothed problem \eqref{eq:SGEP-smooth-Approx}.
Then $0\leq f(\mathbf{x}^{\star})-f(\mathbf{x}_{\epsilon}^{\star})\leq\rho n\left(g_{p}(\epsilon)-\frac{g_{p}^{\prime}(\epsilon)}{2}\epsilon\right)$
and \textup{$\lim_{\epsilon\downarrow0}\rho n\left(g_{p}(\epsilon)-\frac{g_{p}^{\prime}(\epsilon)}{2}\epsilon\right)=0$.}\end{prop}
\begin{IEEEproof}
See Appendix \ref{sec:Proof-of-Proposition_suboptimal}.
\end{IEEEproof}
Proposition \ref{prop:suboptimal} gives a suboptimality bound on
the solution of the smoothed problem \eqref{eq:SGEP-smooth-Approx}
in the sense that we can solve the original problem \eqref{eq:SGEP-Approx}
to a very high accuracy by solving the smoothed problem \eqref{eq:SGEP-smooth-Approx}
with a small enough $\epsilon$ (say $\epsilon\leq10^{-6}$, and $\epsilon=10^{-8}$
is used in our simulations). Of course, in general it is hard to solve
either problem \eqref{eq:SGEP-Approx} or \eqref{eq:SGEP-smooth-Approx}
to global maximum or even local maximum, since both of them are nonconvex.
But from this point, there may be advantages in solving the smoothed
problem with the smoothing parameter $\epsilon$ decreasing gradually,
since choosing a relatively large $\epsilon$ at the beginning can
probably smoothen out some undesirable local maxima, so that the algorithm
can escape from these local points. This idea has been used with some
success in \cite{chartrand2008IRLS,becker2011NESTA}. A decreasing
scheme of $\epsilon$ will be considered later in the numerical simulations.

\subsection{Iteratively Reweighted Quadratic Minorization}

With the quadratic minorization function constructed and the smoothing
technique used to deal with the singularity issue, we are now ready
to state the overall algorithm for the approximate SGEP in \eqref{eq:SGEP-Approx}.

First, we approximate the non-differentiable surrogate functions $g_{p}(x_{i})$
by smooth functions $g_{p}^{\epsilon}(x_{i})$, which leads to the
smoothed problem \eqref{eq:SGEP-smooth-Approx}. Then at iteration
$k$, we construct the quadratic minorization function $\mathbf{x}^{T}\left(\mathbf{A}-\rho{\rm Diag}(\mathbf{w}^{(k)})\right)\mathbf{x}-\rho\sum_{i=1}^{n}c_{i}^{(k)}$
of the objective via majorizing each smoothed surrogate function $g_{p}^{\epsilon}(x_{i})$
at $x_{i}^{(k)}$ by a quadratic function $w_{i}^{(k)}x_{i}^{2}+c_{i}^{(k)}$
and solve the following minorized problem (with the constant term
ignored) 
\begin{equation}
\begin{array}{ll}
\underset{\mathbf{x}}{\mathsf{maximize}} & \mathbf{x}^{T}\left(\mathbf{A}-\rho{\rm Diag}(\mathbf{w}^{(k)})\right)\mathbf{x}\\
\mathsf{subject\; to} & \mathbf{x}^{T}\mathbf{B}\mathbf{x}=1,
\end{array}\label{eq:SGEP-minorized}
\end{equation}
which is to find the leading generalized eigenvector of the matrix
pair $(\mathbf{A}-\rho{\rm Diag}(\mathbf{w}^{(k)}),\mathbf{B}),$
where $\mathbf{w}^{(k)}=[w_{1}^{(k)},\ldots,w_{n}^{(k)}]^{T}$ and
$w_{i}^{(k)},i=1,\ldots,n$ are given in Table \ref{tab:Smooth-surrogate-functions}.
The method is summarized in Algorithm \ref{alg:MM-SGEP} and we will
refer to it as IRQM (Iterative Reweighed Quadratic Minorization),
since it is based on iteratively minorizing the penalty function with
reweighted quadratic function.

\begin{algorithm}[tbh]
\begin{algor}[1]
\item [{Require:}] $\mathbf{A}\in\mathbf{S}^{n}$, $\mathbf{B}\in\mathbf{S}_{++}^{n}$,
$\rho>0$, $\epsilon>0$
\item [{{*}}] Set $k=0$, choose $\mathbf{x}^{(0)}\in\{\mathbf{x}:\mathbf{x}^{T}\mathbf{B}\mathbf{x}=1\}$
\item [{repeat}]~

\begin{algor}[1]
\item [{{*}}] Compute $\mathbf{w}^{(k)}$ according to Table \ref{tab:Smooth-surrogate-functions}.
\item [{{*}}] $\mathbf{x}^{(k+1)}\leftarrow$leading generalized eigenvector
of the matrix pair $(\mathbf{A}-\rho{\rm Diag}(\mathbf{w}^{(k)}),\mathbf{B})$
\item [{{*}}] $k\leftarrow k+1$
\end{algor}
\item [{until}] convergence
\item [{{*}}] \textbf{return} $\mathbf{x}^{(k)}$
\end{algor}
\protect\caption{\label{alg:MM-SGEP}IRQM - Iteratively Reweighed Quadratic Minorization
algorithm for the sparse generalized eigenvalue problem \eqref{eq:SGEP-Approx}.}
\end{algorithm}

At every iteration of the proposed IRQM algorithm, we need to find
the generalized eigenvector of the matrix pair $(\mathbf{A}-\rho{\rm Diag}(\mathbf{w}^{(k)}),\mathbf{B})$
corresponding to the largest generalized eigenvalue. Since $\mathbf{B}\in\mathbf{S}_{++}^{n},$
a standard approach for this problem is to transform it to a standard
eigenvalue problem via the Cholesky decomposition of $\mathbf{B}$.
Then standard algorithms, such as power iterations (applied to a shifted
matrix) and Lanczos method can be used. The drawback of this approach
is that a matrix factorization is needed, making it less attractive
when this factorization is expensive. Besides, as some $w_{i}^{(k)}$
become very large, the problem is highly ill-conditioned and standard
iterative algorithms may suffer from extremely slow convergence.

To overcome these difficulties, we provide a preconditioned steepest
ascent method, which is matrix factorization free and employes preconditioning
to deal with the ill-conditioning problem. Let us derive the steepest
ascent method without preconditioning first. The key step is to reformulate
the leading generalized eigenvalue problem as maximizing the Rayleigh
quotient
\begin{equation}
R(\mathbf{x})=\frac{\mathbf{x}^{T}\tilde{\mathbf{A}}\mathbf{x}}{\mathbf{x}^{T}\mathbf{B}\mathbf{x}}
\end{equation}
over the domain $\mathbf{x}\neq\mathbf{0}$, where $\tilde{\mathbf{A}}=\mathbf{A}-\rho{\rm Diag}(\mathbf{w}^{(k)}).$
Let $\mathbf{x}^{(l)}$ be the current iterate, the gradient of $R(\mathbf{x})$
at $\mathbf{x}^{(l)}$ is 
\[
\frac{2}{\mathbf{x}^{(l)T}\mathbf{B}\mathbf{x}^{(l)}}\left(\tilde{\mathbf{A}}\mathbf{x}^{(l)}-R(\mathbf{x}^{(l)})\mathbf{B}\mathbf{x}^{(l)}\right),
\]
which is an ascent direction of $R(\mathbf{x})$ at $\mathbf{x}^{(l)}$.
Let $\mathbf{r}^{(l)}=\tilde{\mathbf{A}}\mathbf{x}^{(l)}-R(\mathbf{x}^{(l)})\mathbf{B}\mathbf{x}^{(l)}$,
the steepest ascent method searches along the line $\mathbf{x}^{(l)}+\tau\mathbf{r}^{(l)}$
for a $\tau$ that maximizes the Rayleigh quotient $R(\mathbf{x}^{(l)}+\tau\mathbf{r}^{(l)})$.
Since the Rayleigh quotient $R(\mathbf{x}^{(l)}+\tau\mathbf{r}^{(l)})$
is a scalar function of $\tau$, the maximum will be achieved either
at points with zero derivative or as $\tau$ goes to infinity. Setting
the derivative of $R(\mathbf{x}^{(l)}+\tau\mathbf{r}^{(l)})$ with
respect to $\tau$ equal to 0, we can get the following quadratic
equation
\begin{equation}
a\tau^{2}+b\tau+c=0,\label{eq:quad_equation-1}
\end{equation}
where 
\begin{eqnarray*}
a & = & \mathbf{r}^{(l)T}\tilde{\mathbf{A}}\mathbf{r}^{(l)}\mathbf{x}^{(l)T}\mathbf{B}\mathbf{r}^{(l)}-\mathbf{r}^{(l)T}\mathbf{B}\mathbf{r}^{(l)}\mathbf{x}^{(l)T}\tilde{\mathbf{A}}\mathbf{r}^{(l)}\\
b & = & \mathbf{r}^{(l)T}\tilde{\mathbf{A}}\mathbf{r}^{(l)}\mathbf{x}^{(l)T}\mathbf{B}\mathbf{x}^{(l)}-\mathbf{r}^{(l)T}\mathbf{B}\mathbf{r}^{(l)}\mathbf{x}^{(l)T}\tilde{\mathbf{A}}\mathbf{x}^{(l)}\\
c & = & \mathbf{r}^{(l)T}\tilde{\mathbf{A}}\mathbf{x}^{(l)}\mathbf{x}^{(l)T}\mathbf{B}\mathbf{x}^{(l)}-\mathbf{r}^{(l)T}\mathbf{B}\mathbf{x}^{(l)}\mathbf{x}^{(l)T}\tilde{\mathbf{A}}\mathbf{x}^{(l)}.
\end{eqnarray*}
Let us denote $B_{{\rm xx}}=\mathbf{x}^{(l)T}\mathbf{B}\mathbf{x}^{(l)}$,
$B_{{\rm rr}}=\mathbf{r}^{(l)T}\mathbf{B}\mathbf{r}^{(l)}$ and $B_{{\rm xr}}=\mathbf{x}^{(l)T}\mathbf{B}\mathbf{r}^{(l)}$,
by direct computation we have
\[
\begin{aligned}b^{2}-4ac= & \left(B_{{\rm xx}}B_{{\rm rr}}\big(R(\mathbf{r}^{(l)})-R(\mathbf{x}^{(l)})\big)-2B_{{\rm xr}}\left\Vert \mathbf{r}^{(l)}\right\Vert _{2}^{2}\right)^{2}\\
 & +4\left\Vert \mathbf{r}^{(l)}\right\Vert _{2}^{4}\left(B_{{\rm xx}}B_{{\rm rr}}-B_{{\rm xr}}^{2}\right).
\end{aligned}
\]
According to Cauchy-Schwartz inequality, it is easy to see that $B_{{\rm xx}}B_{{\rm rr}}-B_{{\rm xr}}^{2}\geq0$,
thus $b^{2}-4ac\geq0$, which implies that the equation \eqref{eq:quad_equation-1}
has one or two real roots. By comparing the Rayleigh quotient $R(\mathbf{x}^{(l)}+\tau\mathbf{r}^{(l)})$
at the roots of equation \eqref{eq:quad_equation-1} with $R(\mathbf{r}^{(l)})$
(the Rayleigh quotient corresponding to $\tau\rightarrow\infty$),
we can determine the steepest ascent. It is worth noting that the
coefficients of the equation \eqref{eq:quad_equation-1} can be computed
by matrix-vector multiplications and inner products only, thus very
efficient. 

Though the per-iteration computational complexity of this steepest
ascent method is very low, it may converge very slow, especially when
some $w_{i}^{(k)}$ become very large. To accelerate the convergence,
we introduce a preconditioner here. Preconditioning is an important
technique in iterative methods for solving large system of linear
equations, for example the widely used preconditioned conjugate gradient
method. It can also be used for eigenvalue problems. In the steepest
ascent method, to introduce a positive definite preconditioner $\mathbf{P}$,
we simply multiply the residual $\mathbf{r}^{(l)}$ by $\mathbf{P}$.
The steepest ascent method with preconditioning for the leading generalized
eigenvalue problem \eqref{eq:SGEP-minorized} is summarized in Algorithm
\ref{alg:steepest_ascent-1}. To use the algorithm in practice, the
preconditioner $\mathbf{P}$ remains to be chosen. For the particular
problem of interest, we choose a diagonal $\mathbf{P}$ as follows
\[
\mathbf{P}=\begin{cases}
{\rm Diag}\left(\text{\ensuremath{\rho}}\mathbf{w}^{(k)}+\left|{\rm diag}(\mathbf{A})\right|\right)^{-1}, & \frac{\text{\ensuremath{\rho}}\left\Vert \mathbf{w}^{(k)}\right\Vert _{2}}{\left\Vert {\rm diag}(\mathbf{A})\right\Vert _{2}}>10^{2}\\
\mathbf{I}_{n}. & {\rm otherwise.}
\end{cases}
\]
In other words, we apply a preconditioner only when some elements
of $\mathbf{w}^{(k)}$ become relatively large. Since the preconditioner
$\mathbf{P}$ we choose here is positive definite, the direction $\mathbf{P}\mathbf{r}^{(l)}$
is still an ascent direction and the algorithm is still monotonically
increasing. For more details regarding preconditioned eigensolvers,
the readers can refer to the book \cite{EigBook}.

In practice, the preconditioned steepest ascent method usually converges
to the leading generalized eigenvector, but it is not guaranteed in
principle\textcolor{black}{, since the }Rayleigh quotient\textcolor{black}{{}
is not concave. But note that the descent property \eqref{eq:descent-property}
of the majorization-minimization scheme depends only on decreasing
$u(\mathbf{x},\mathbf{x}^{(k)})$ and not on minimizing it. Similarly,
for the minorization-maximization scheme used by Algorithm \ref{alg:MM-SGEP},
to preserve the ascent property, we only need to increase the objective
of \eqref{eq:SGEP-minorized} at each iteration, rather than maximizing
it. Since the steepest ascent method increases the objective at every
iteration, thus when it is applied (initialized with the solution
of previous iteration) to compute the leading generalized eigenvector
at each iteration of Algorithm \ref{alg:MM-SGEP}, the ascent property
of Algorithm \ref{alg:MM-SGEP} can be guaranteed.} 

\begin{algorithm}[tbh]
\begin{algor}[1]
\item [{Require:}] $\mathbf{A}\in\mathbf{S}^{n}$, $\mathbf{B}\in\mathbf{S}_{++}^{n}$,
$\mathbf{w}^{(k)}$, $\rho>0$
\item [{{*}}] Set $l=0$, choose $\mathbf{x}^{(0)}\in\{\mathbf{x}:\mathbf{x}^{T}\mathbf{B}\mathbf{x}=1\}$
\item [{{*}}] Let $\tilde{\mathbf{A}}=\mathbf{A}-\rho{\rm Diag}(\mathbf{w}^{(k)})$
\item [{repeat}]~

\begin{algor}[1]
\item [{{*}}] $R(\mathbf{x}^{(l)})=\mathbf{x}^{(l)T}\tilde{\mathbf{A}}\mathbf{x}^{(l)}/\mathbf{x}^{(l)T}\mathbf{B}\mathbf{x}^{(l)}$
\item [{{*}}] $\mathbf{r}^{(l)}=\tilde{\mathbf{A}}\mathbf{x}^{(l)}-R(\mathbf{x}^{(l)})\mathbf{B}\mathbf{x}^{(l)}$
\item [{{*}}] $\mathbf{r}^{(l)}=\mathbf{P}\mathbf{r}^{(l)}$ 
\item [{{*}}] $\mathbf{x}=\mathbf{x}^{(l)}+\tau\mathbf{r}^{(l)}$, with
$\tau$ chosen to maximize $R(\mathbf{x}^{(l)}+\tau\mathbf{r}^{(l)})$
\item [{{*}}] $\mathbf{x}^{(l+1)}=\mathbf{x}/\sqrt{\mathbf{x}^{T}\mathbf{B}\mathbf{x}}$
\item [{{*}}] $l=l+1$
\end{algor}
\item [{until}] convergence
\item [{{*}}] \textbf{return} $\mathbf{x}^{(l)}$
\end{algor}
\protect\caption{\label{alg:steepest_ascent-1}Preconditioned steepest ascent method
for problem \eqref{eq:SGEP-minorized}.}
\end{algorithm}

\subsection{Sparse GEP with Special Structure}

Until now we have considered the sparse GEP in the general case with
$\mathbf{A}\in\mathbf{S}^{n}$, $\mathbf{B}\in\mathbf{S}_{++}^{n}$
and derived an iterative algorithm IRQM. If we assume more properties
or some special structure for $\mathbf{A}$ and $\mathbf{B}$, then
we may derive simpler and more efficient algorithms. In the following,
we will consider the case where $\mathbf{A}\in\mathbf{S}_{+}^{n}$
and $\mathbf{B}={\rm Diag}(\mathbf{b})$, $\mathbf{b}\in\mathbf{R}_{++}^{n}$.
Notice that although this is a special case of the general sparse
GEP, it still includes the sparse PCA problem as a special case where
$\mathbf{B}=\mathbf{I}_{n}$.\emph{}

We first present two results that will be used when deriving fast
algorithms for this special case.
\begin{prop}
\label{prop:nonconvex-QCQP}Given $\mathbf{a}\in\mathbf{R}^{n},$
$\mathbf{w},\mathbf{b}\in\mathbf{R}_{++}^{n}$, $\rho>0$, let $\mathcal{I}_{{\rm min}}=\arg\min\{\rho w_{i}/b_{i}:i\in\{1,\ldots,n\}\}$,
\textup{$\mu_{{\rm min}}=-\min\{\rho w_{i}/b_{i}:i\in\{1,\ldots,n\}\}$
and $s=\sum_{i\notin\mathcal{I}_{{\rm min}}}\frac{b_{i}a_{i}^{2}}{(\mu_{{\rm min}}b_{i}+\rho w_{i})^{2}}$.
Then the problem} 
\begin{equation}
\begin{array}{ll}
\underset{\mathbf{x}}{\mathsf{maximize}} & 2\mathbf{a}^{T}\mathbf{x}-\rho\mathbf{x}^{T}{\rm Diag}(\mathbf{w})\mathbf{x}\\
\mathsf{subject\; to} & \mathbf{x}^{T}{\rm Diag}(\mathbf{b})\mathbf{x}=1
\end{array}\label{eq:non-convex-QCQP-2}
\end{equation}
admits the following solution:
\begin{itemize}
\item If $\exists i\in\mathcal{I}_{{\rm min}}$, such that $a_{i}^{2}>0$
or $s>1$, then 
\[
x_{i}^{\star}=\frac{a_{i}}{\mu b_{i}+\rho w_{i}},\, i=1,\ldots,n,
\]
where $\mu>\mu_{{\rm min}}$ is given by the solution of the scalar
equation
\[
\sum_{i=1}^{n}\frac{b_{i}a_{i}^{2}}{(\mu b_{i}+\rho w_{i})^{2}}=1.
\]

\item Otherwise,
\[
x_{i}^{\star}=\begin{cases}
a_{i}/\left(\mu_{{\rm min}}b_{i}+\rho w_{i}\right), & i\notin\mathcal{I}_{{\rm min}}\\
\sqrt{\left(1-s\right)/b_{i}}, & i=\max\{i:i\in\mathcal{I}_{{\rm min}}\}\\
0, & {\rm otherwise}.
\end{cases}
\]

\end{itemize}
\end{prop}
\begin{IEEEproof}
See Appendix \ref{sec:Proof-of-Proposition_QCQP}.\end{IEEEproof}
\begin{prop}
\label{prop:L0-closed-form}Given $\mathbf{a}\in\mathbf{R}^{n}$ with
$\left|a_{1}\right|\geq\ldots\geq\left|a_{n}\right|$ and $\rho>0$,
then the problem 
\begin{equation}
\begin{array}{ll}
\underset{\mathbf{x}}{\mathsf{maximize}} & \mathbf{a}^{T}\mathbf{x}-\rho\left\Vert \mathbf{x}\right\Vert _{0}\\
\mathsf{subject\; to} & \left\Vert \mathbf{x}\right\Vert _{2}=1
\end{array}\label{eq:L0-penalized}
\end{equation}
admits the following solution:
\begin{itemize}
\item If $\left|a_{1}\right|\leq\rho,$ then 
\begin{equation}
x_{i}^{\star}=\begin{cases}
{\rm sgn}(a_{1}), & i=1\\
0, & {\rm otherwise}.
\end{cases}\label{eq:a_less_rho}
\end{equation}

\item Otherwise,
\[
x_{i}^{\star}=\begin{cases}
a_{i}/\sqrt{\sum_{j=1}^{s}a_{j}^{2}}, & i\leq s\\
0, & {\rm otherwise},
\end{cases}
\]
where $s$ is the largest integer $p$ that satisfies the following
inequality
\begin{equation}
\sqrt{{\textstyle \sum_{i=1}^{p}}a_{i}^{2}}>\sqrt{{\textstyle \sum_{i=1}^{p-1}}a_{i}^{2}}+\rho.\label{eq:L0-inequal-1-1}
\end{equation}

\end{itemize}
\end{prop}
\begin{IEEEproof}
See Appendix \ref{sec:Proof-of-Proposition_L0_closed_form}.
\end{IEEEproof}
Let us return to the problem. In this special case, the smoothed problem
\eqref{eq:SGEP-smooth-Approx} reduces to 
\begin{equation}
\begin{array}{ll}
\underset{\mathbf{x}}{\mathsf{maximize}} & \mathbf{x}^{T}\mathbf{A}\mathbf{x}-\rho\sum_{i=1}^{n}g_{p}^{\epsilon}(x_{i})\\
\mathsf{subject\; to} & \mathbf{x}^{T}{\rm Diag}(\mathbf{b})\mathbf{x}=1.
\end{array}\label{eq:diagonal-approx-penalized}
\end{equation}
The previously derived IRQM algorithm can be used here, but in that
iterative algorithm, we need to find the leading generalized eigenvector
at each iteration, for which another iterative algorithm is needed.
By exploiting the special structure of this case, in the following
we derive a simpler algorithm that at each iteration has a closed-form
solution. 

Notice that, in this case, $\mathbf{A}\in\mathbf{S}_{+}^{n}$, the
first term $\mathbf{x}^{T}\mathbf{A}\mathbf{x}$ in the objective
is convex and can be minorized by its tangent plane $2\mathbf{x}^{T}\mathbf{A}\mathbf{x}^{(k)}$
at $\mathbf{x}^{(k)}$. So instead of only minorizing the second term,
we can minorize both terms. This suggests solving the following minorized
problem at iteration $k$:
\begin{equation}
\begin{array}{ll}
\underset{\mathbf{x}}{\mathsf{maximize}} & 2\mathbf{x}^{T}\mathbf{A}\mathbf{x}^{(k)}-\rho\mathbf{x}^{T}{\rm Diag}(\mathbf{w}^{(k)})\mathbf{x}\\
\mathsf{subject\; to} & \mathbf{x}^{T}{\rm Diag}(\mathbf{b})\mathbf{x}=1,
\end{array}\label{eq:non-convex-QCQP-1}
\end{equation}
where $\mathbf{x}^{(k)}$ is the solution at iteration $k$ and $\mathbf{w}^{(k)}$
is computed according to Table \ref{tab:Smooth-surrogate-functions}.
The problem is a nonconvex QCQP, but by letting $\mathbf{a}=\mathbf{A}\mathbf{x}^{(k)}$
and $\mathbf{w}=\mathbf{w}^{(k)}$, we know from Proposition \ref{prop:nonconvex-QCQP}
that it can be solved in closed-form. The iterative algorithm for
solving problem \eqref{eq:diagonal-approx-penalized} is summarized
in Algorithm \ref{alg:approx-penalized-diagonal}.

\begin{algorithm}[tbh]
\begin{algor}[1]
\item [{Require:}] $\mathbf{A}\in\mathbf{S}_{+}^{n}$, $\mathbf{b}\in\mathbf{R}_{++}^{n}$,
$\rho>0$, $\epsilon>0$
\item [{{*}}] Set $k=0$, choose $\mathbf{x}^{(0)}\in\{\mathbf{x}:\mathbf{x}^{T}{\rm Diag}(\mathbf{b})\mathbf{x}=1\}$
\item [{repeat}]~

\begin{algor}[1]
\item [{{*}}] $\mathbf{a}=\mathbf{A}\mathbf{x}^{(k)}$
\item [{{*}}] Compute $\mathbf{w}^{(k)}$ according to Table \ref{tab:Smooth-surrogate-functions}.
\item [{{*}}] Solve the following problem according to Proposition \ref{prop:nonconvex-QCQP}
and set the solution as $\mathbf{x}^{(k+1)}$: 
\[
\max_{\mathbf{x}}\{2\mathbf{a}^{T}\mathbf{x}-\rho\mathbf{x}^{T}{\rm Diag}(\mathbf{w}^{(k)})\mathbf{x}:\mathbf{x}^{T}{\rm Diag}(\mathbf{b})\mathbf{x}=1\}
\]

\item [{{*}}] $k=k+1$
\end{algor}
\item [{until}] convergence
\item [{{*}}] \textbf{return} $\mathbf{x}^{(k)}$
\end{algor}
\protect\caption{\label{alg:approx-penalized-diagonal}The MM algorithm for problem
\eqref{eq:diagonal-approx-penalized}.}
\end{algorithm}

In fact, in this special case, we can apply the MM scheme to solve
the original problem \eqref{eq:SGEP-L0} directly, without approximating
$\left\Vert \mathbf{x}\right\Vert _{0},$ i.e., solving
\begin{equation}
\begin{array}{ll}
\underset{\mathbf{x}}{\mathsf{maximize}} & \mathbf{x}^{T}\mathbf{A}\mathbf{x}-\rho\left\Vert \mathbf{x}\right\Vert _{0}\\
\mathsf{subject\; to} & \mathbf{x}^{T}{\rm Diag}(\mathbf{b})\mathbf{x}=1.
\end{array}\label{eq:L0-penalized-diagonal}
\end{equation}

First, we define a new variable $\tilde{\mathbf{x}}={\rm Diag}(\mathbf{b})^{\frac{1}{2}}\mathbf{x}$
and using the fact $\left\Vert {\rm Diag}(\mathbf{b})^{-\frac{1}{2}}\tilde{\mathbf{x}}\right\Vert _{0}=\left\Vert \tilde{\mathbf{x}}\right\Vert _{0}$,
the problem can be rewritten as
\begin{equation}
\begin{array}{ll}
\underset{\tilde{\mathbf{x}}}{\mathsf{maximize}} & \tilde{\mathbf{x}}^{T}\tilde{\mathbf{A}}\tilde{\mathbf{x}}-\rho\left\Vert \tilde{\mathbf{x}}\right\Vert _{0}\\
\mathsf{subject\; to} & \tilde{\mathbf{x}}^{T}\tilde{\mathbf{x}}=1,
\end{array}
\end{equation}
where $\tilde{\mathbf{A}}={\rm Diag}(\mathbf{b})^{-\frac{1}{2}}\mathbf{A}{\rm Diag}(\mathbf{b})^{-\frac{1}{2}}$.

Now the idea is to minorize only the quadratic term by its tangent
plane, while keeping the $\ell_{0}$-norm. Given $\tilde{\mathbf{x}}^{(k)}$
at iteration $k$, linearizing the quadratic term yields
\begin{equation}
\begin{array}{ll}
\underset{\tilde{\mathbf{x}}}{\mathsf{maximize}} & 2\tilde{\mathbf{x}}^{T}\tilde{\mathbf{A}}\tilde{\mathbf{x}}^{(k)}-\rho\left\Vert \tilde{\mathbf{x}}\right\Vert _{0}\\
\mathsf{subject\; to} & \left\Vert \tilde{\mathbf{x}}\right\Vert _{2}=1,
\end{array}
\end{equation}
which has a closed-form solution. To see this, we first define $\mathbf{a}=2\tilde{\mathbf{A}}\tilde{\mathbf{x}}^{(k)}$
and sort the entries of vector $\mathbf{a}$ according to the absolute
value (only needed for entries with $\left|a_{i}\right|>\rho$) in
descending order, then Proposition \ref{prop:L0-closed-form} can
be readily applied to obtain the solution. Finally we need to reorder
the solution back to the original ordering. This algorithm for solving
problem \eqref{eq:L0-penalized-diagonal} is summarized in Algorithm
\ref{alg:L0-penalized-diagonal}.  \textbf{}

It is worth noting that although the derivations of Algorithms \ref{alg:approx-penalized-diagonal}
and \ref{alg:L0-penalized-diagonal} require $\mathbf{A}$ to be symmetric
positive semidefinite, the algorithms can also be used to deal with
the more general case $\mathbf{A}\in\mathbf{S}^{n}$. When the matrix
$\mathbf{A}$ in problem \eqref{eq:diagonal-approx-penalized} or
\eqref{eq:L0-penalized-diagonal} is not positive semidefinite, we
can replace $\mathbf{A}$ with $\mathbf{A}_{\alpha}=\mathbf{A}+\alpha{\rm Diag}(\mathbf{b}),$
with $\alpha\geq-\lambda_{{\rm min}}(\mathbf{A})/b_{{\rm min}}$ such
that $\mathbf{A}_{\alpha}\in\mathbf{S}_{+}^{n},$ where $\lambda_{{\rm min}}(\mathbf{A})$
is the smallest eigenvalue of matrix $\mathbf{A}$ and $b_{{\rm min}}$
is the smallest entry of $\mathbf{b}$. Since the additional term
$\alpha\mathbf{x}^{T}{\rm Diag}(\mathbf{b})\mathbf{x}$ in the objective
is just a constant $\alpha$ over the constraint set, it is easy to
see that after replacing $\mathbf{A}$ with $\mathbf{A}_{\alpha}$
the resulting problem is equivalent to the original one. Then the
Algorithm \ref{alg:approx-penalized-diagonal} or \ref{alg:L0-penalized-diagonal}
can be readily applied.

\begin{algorithm}[tbh]
\begin{algor}[1]
\item [{Require:}] $\mathbf{A}\in\mathbf{S}_{+}^{n}$, $\mathbf{b}\in\mathbf{R}_{++}^{n}$,
$\rho>0$
\item [{{*}}] Set $k=0$, choose $\tilde{\mathbf{x}}^{(0)}\in\{\tilde{\mathbf{x}}:\tilde{\mathbf{x}}^{T}\tilde{\mathbf{x}}=1\}$
\item [{{*}}] Let $\tilde{\mathbf{A}}={\rm Diag}(\mathbf{b})^{-\frac{1}{2}}\mathbf{A}{\rm Diag}(\mathbf{b})^{-\frac{1}{2}}$
\item [{repeat}]~

\begin{algor}[1]
\item [{{*}}] $\mathbf{a}=2\tilde{\mathbf{A}}\tilde{\mathbf{x}}^{(k)}$
\item [{{*}}] \begin{raggedright}
Sort $\mathbf{a}$ with the absolute value in descending
order. 
\par\end{raggedright}
\item [{{*}}] Compute $\tilde{\mathbf{x}}^{(k+1)}$ according to Proposition
\ref{prop:L0-closed-form}:
\[
\tilde{\mathbf{x}}^{(k+1)}\in\arg\max_{\mathbf{\tilde{x}}}\{\mathbf{a}^{T}\tilde{\mathbf{x}}-\rho\left\Vert \tilde{\mathbf{x}}\right\Vert _{0}:\left\Vert \tilde{\mathbf{x}}\right\Vert _{2}=1\}
\]

\item [{{*}}] Reorder $\tilde{\mathbf{x}}^{(k+1)}$ 
\item [{{*}}] $k=k+1$
\end{algor}
\item [{until}] convergence
\item [{{*}}] \textbf{return} $\mathbf{x}={\rm Diag}(\mathbf{b})^{-\frac{1}{2}}\tilde{\mathbf{x}}^{(k)}$
\end{algor}
\protect\caption{\label{alg:L0-penalized-diagonal}The MM algorithm for problem \eqref{eq:L0-penalized-diagonal}.}
\end{algorithm}

\section{Convergence analysis\label{sec:Convergence-analysis}}

The algorithms proposed in this paper are all based on the minorization-maximization
scheme, thus according to subsection \ref{sub:MM-method}, we know
that the sequence of objective values evaluated at $\{\mathbf{x}^{(k)}\}$
generated by the algorithms is non-decreasing. Since the constraint
sets in our problems are compact, the sequence of objective values
is bounded. Thus, the sequence of objective values is guaranteed to
converge to a finite value. The monotonicity makes MM algorithms very
stable. In this section, we will analyze the convergence property
of the sequence $\{\mathbf{x}^{(k)}\}$ generated by the algorithms. 

Let us consider the IRQM algorithm in Algorithm \ref{alg:MM-SGEP},
in which the minorization-maximization scheme is applied to the smoothed
problem \eqref{eq:SGEP-smooth-Approx}. In the problem, the objective
is neither convex nor concave and the constraint set is also nonconvex.
But as we shall see later, after introducing a technical assumption
on the surrogate function $g_{p}(x)$, the problem is equivalent to
a problem which maximizes a convex function over a convex set and
we will prove that the sequence generated by the IRQM algorithm converges
to the stationary point of the equivalent problem. The convergence
of the Algorithm \ref{alg:approx-penalized-diagonal} can be proved
similarly, since the minorization function applied can also be convexified.
First, let us give the assumption and present some results that will
be useful later.

\begin{assumption}\label{assump}The surrogate function $g_{p}(x)$
is twice differentiable on $(0,+\infty)$ and its gradient $g_{p}^{\prime}(x)$
is convex on $(0,+\infty)$.\end{assumption}

It is easy to verify that the three surrogate functions listed in
Table \eqref{tab:Smooth-surrogate-functions} all satisfy this assumption.
With this assumption, the first result shows that the smooth approximation
$g_{p}^{\epsilon}(x)$ we have constructed is Lipschitz continuously
differentiable.
\begin{lem}
\label{lem:Lipschitz_gradient}Let $g_{p}(\cdot)$ be a continuous
even function defined on $\mathbf{R}$, differentiable everywhere
except at zero, concave and monotone increasing on $[0,+\infty)$
with $g_{p}(0)=0.$ Let Assumption \ref{assump} be satisfied. Then
the smooth approximation $g_{p}^{\epsilon}(x)$ defined by \eqref{eq:smooth-surrogate}
is Lipschitz continuously differentiable with Lipschitz constant $L=\max\{\frac{g_{p}^{\prime}(\epsilon)}{\epsilon},\left|g_{p}^{\prime\prime}(\epsilon)\right|\}.$\end{lem}
\begin{IEEEproof}
See Appendix \ref{sub:Proof-of-Lemma_Lipschitz}.
\end{IEEEproof}
Next, we recall a useful property of Lipschitz continuously differentiable
functions \cite{zlobec2005liu}. 
\begin{prop}
\label{prop:convexify}If $f:\mathbf{R}^{n}\rightarrow\mathbf{R}$
is Lipschitz continuously differentiable on a convex set $C$ with
some Lipschitz constant $L$, then $\varphi(\mathbf{x})=f(\mathbf{x})+\frac{\alpha}{2}\mathbf{x}^{T}\mathbf{x}$
is a convex function on $C$ for every $\alpha\geq L.$
\end{prop}
The next result then follows, showing that the smoothed problem \eqref{eq:SGEP-smooth-Approx}
is equivalent to a problem in the form of maximizing a convex function
over a compact set.
\begin{lem}
\label{lem:obj-convex}There exists $\alpha>0$ such that $\mathbf{x}^{T}\left(\mathbf{A}+\alpha\mathbf{B}\right)\mathbf{x}-\rho\sum_{i=1}^{n}g_{p}^{\epsilon}(x_{i})$
is convex and the problem 
\begin{equation}
\begin{array}{ll}
\underset{\mathbf{x}}{\mathsf{maximize}} & \mathbf{x}^{T}\left(\mathbf{A}+\alpha\mathbf{B}\right)\mathbf{x}-\rho\sum_{i=1}^{n}g_{p}^{\epsilon}(x_{i})\\
\mathsf{subject\; to} & \mathbf{x}^{T}\mathbf{B}\mathbf{x}=1
\end{array}\label{eq:convex-obj-problem}
\end{equation}
is equivalent to the problem \eqref{eq:SGEP-smooth-Approx} in the
sense that they admit the same set of optimal solutions.\end{lem}
\begin{IEEEproof}
From Lemma \ref{lem:Lipschitz_gradient}, it is easy to see that $\mathbf{x}^{T}\mathbf{A}\mathbf{x}-\rho\sum_{i=1}^{n}g_{p}^{\epsilon}(x_{i})$
is Lipschitz continuously differentiable. Assume the Lipschitz constant
of its gradient is $L$, then according to Proposition \ref{prop:convexify},
$\mathbf{x}^{T}\left(\mathbf{A}+\frac{L}{2}\mathbf{I}\right)\mathbf{x}-\rho\sum_{i=1}^{n}g_{p}^{\epsilon}(x_{i})$
is convex. Since $\mathbf{B}$ is positive definite, $\lambda_{{\rm min}}(\mathbf{B})>0$.
By choosing $\alpha\geq\frac{L}{2\lambda_{{\rm min}}(\mathbf{B})},$
we have that $\mathbf{x}^{T}(\alpha\mathbf{B}-\frac{L}{2}\mathbf{I})\mathbf{x}$
is convex. The sum of the two convex functions, i.e., $\mathbf{x}^{T}\left(\mathbf{A}+\alpha\mathbf{B}\right)\mathbf{x}-\rho\sum_{i=1}^{n}g_{p}^{\epsilon}(x_{i})$,
is convex.

Since the additional term $\alpha\mathbf{x}^{T}\mathbf{B}\mathbf{x}$
is just a constant $\alpha$ over the constraint set $\mathbf{x}^{T}\mathbf{B}\mathbf{x}=1$,
it is obvious that any solution of problem \eqref{eq:convex-obj-problem}
is also a solution of problem \eqref{eq:SGEP-smooth-Approx} and vice
versa.
\end{IEEEproof}
Generally speaking, maximizing a convex function over a compact set
remains a hard nonconvex problem. There is some consolation, however,
according to the following result in convex analysis \cite{rockafellar1997convex}.
\begin{prop}
\label{prop:max-convex}Let $f:\mathbf{R}^{n}\rightarrow\mathbf{R}$
be a convex function. Let $S\in\mathbf{R}^{n}$ be an arbitrary set
and ${\rm conv}(S)$ be its convex hull. Then 
\[
\sup\{f(\mathbf{x})|\mathbf{x}\in{\rm conv}(S)\}=\sup\{f(\mathbf{x})|\mathbf{x}\in S\},
\]
where the first supremum is attained only when the second (more restrictive)
supremum is attained.
\end{prop}
According to Proposition \ref{prop:max-convex}, we can further relax
the constraint $\mathbf{x}^{T}\mathbf{B}\mathbf{x}=1$ in problem
\eqref{eq:convex-obj-problem} to $\mathbf{x}^{T}\mathbf{B}\mathbf{x}\leq1$,
namely, the problem 
\begin{equation}
\begin{array}{ll}
\underset{\mathbf{x}}{\mathsf{maximize}} & \mathbf{x}^{T}\left(\mathbf{A}+\alpha\mathbf{B}\right)\mathbf{x}-\rho\sum_{i=1}^{n}g_{p}^{\epsilon}(x_{i})\\
\mathsf{subject\; to} & \mathbf{x}^{T}\mathbf{B}\mathbf{x}\leq1
\end{array}\label{eq:convex-constraint}
\end{equation}
is still equivalent to problem \eqref{eq:SGEP-smooth-Approx} in the
sense that they admit the same set of optimal solutions. 

Let us denote the objective function of problem \eqref{eq:convex-constraint}
by $f_{\epsilon}^{\alpha}(\mathbf{x})$ and define $\mathcal{B}=\{\mathbf{x}\in\mathbf{R}^{n}|\mathbf{x}^{T}\mathbf{B}\mathbf{x}\leq1\}$,
then a point $\mathbf{x^{\star}}$ is referred to as a stationary
point of problem \eqref{eq:convex-constraint} if 
\begin{equation}
\nabla f_{\epsilon}^{\alpha}(\mathbf{x}^{\star})^{T}(\mathbf{x}-\mathbf{x}^{\star})\leq0,\quad\forall\mathbf{x}\in\mathcal{B}.\label{eq:stationary-point}
\end{equation}

\begin{thm}
Let $\{\mathbf{x}^{(k)}\}$ be the sequence generated by the IRQM
algorithm in Algorithm \ref{alg:MM-SGEP}. Then every limit point
of the sequence $\{\mathbf{x}^{(k)}\}$ is a stationary point of the
problem \eqref{eq:convex-constraint}, which is equivalent%
\footnote{The equivalence of the problem \eqref{eq:convex-constraint} and \eqref{eq:SGEP-smooth-Approx}
is in the sense that they have the same set of optimal solutions,
but they may have different stationary points. The convergence to
a stationary point of problem \eqref{eq:convex-constraint} does not
imply the convergence to a stationary point of problem \eqref{eq:SGEP-smooth-Approx}.%
} to the problem \eqref{eq:SGEP-smooth-Approx}. \end{thm}
\begin{IEEEproof}
Denote the objective function of the problem \eqref{eq:SGEP-smooth-Approx}
by $f_{\epsilon}(\mathbf{x})$ and its quadratic minorization function
at $\mathbf{x}^{(k)}$ by $q(\mathbf{x}|\mathbf{x}^{(k)})$, i.e.,
$q(\mathbf{x}|\mathbf{x}^{(k)})=\mathbf{x}^{T}\left(\mathbf{A}-\rho{\rm Diag}(\mathbf{w}^{(k)})\right)\mathbf{x}.$
Denote $\mathcal{S}=\{\mathbf{x}\in\mathbf{R}^{n}|\mathbf{x}^{T}\mathbf{B}\mathbf{x}=1\}$
and $\mathcal{B}=\{\mathbf{x}\in\mathbf{R}^{n}|\mathbf{x}^{T}\mathbf{B}\mathbf{x}\leq1\}$.
According to the general MM framework, we have
\[
f_{\epsilon}(\mathbf{x}^{(k+1)})\geq q(\mathbf{x}^{(k+1)}|\mathbf{x}^{(k)})\geq q(\mathbf{x}^{(k)}|\mathbf{x}^{(k)})=f_{\epsilon}(\mathbf{x}^{(k)}),
\]
which means $\{f_{\epsilon}(\mathbf{x}^{(k)})\}$ is a non-decreasing
sequence. 

Assume that there exists a converging subsequence $\mathbf{x}^{(k_{j})}\rightarrow\mathbf{x}^{\infty},$
then
\[
\begin{aligned}q(\mathbf{x}^{(k_{j+1})}|\mathbf{x}^{(k_{j+1})}) & =f_{\epsilon}(\mathbf{x}^{(k_{j+1})})\geq f_{\epsilon}(\mathbf{x}^{(k_{j}+1)})\\
 & \geq q(\mathbf{x}^{(k_{j}+1)}|\mathbf{x}^{(k_{j})})\geq q(\mathbf{x}|\mathbf{x}^{(k_{j})}),\forall\mathbf{x}\in\mathcal{S}.
\end{aligned}
\]
 Letting $j\rightarrow+\infty,$ we obtain 
\[
q(\mathbf{x}^{\infty}|\mathbf{x}^{\infty})\geq q(\mathbf{x}|\mathbf{x}^{\infty}),\,\forall\mathbf{x}\in\mathcal{S}.
\]

It is easy to see that we can always find $\alpha>0$ such that $q_{\alpha}(\mathbf{x}|\mathbf{x}^{\infty})=q(\mathbf{x}|\mathbf{x}^{\infty})+\alpha\mathbf{x}^{T}\mathbf{B}\mathbf{x}$
is convex and $\mathbf{x}^{\infty}$ is still a global maximizer of
$q_{\alpha}(\mathbf{x}|\mathbf{x}^{\infty})$ over $\mathcal{S}$.
Due to Lemma \ref{lem:obj-convex}, we can always choose $\alpha$
large enough such that $f_{\epsilon}^{\alpha}(\mathbf{x})=f_{\epsilon}(\mathbf{x})+\alpha\mathbf{x}^{T}\mathbf{B}\mathbf{x}$
is also convex. By Proposition \ref{prop:max-convex}, we have 
\[
q_{\alpha}(\mathbf{x}^{\infty}|\mathbf{x}^{\infty})\geq q_{\alpha}(\mathbf{x}|\mathbf{x}^{\infty}),\,\forall\mathbf{x}\in\mathcal{B},
\]
i.e., $\mathbf{x}^{\infty}$ is a global maximizer of $q_{\alpha}(\mathbf{x}|\mathbf{x}^{\infty})$
over the convex set $\mathcal{B}$. As a necessary condition, we get
\[
\nabla q_{\alpha}(\mathbf{x}^{\infty}|\mathbf{x}^{\infty})^{T}(\mathbf{x}-\mathbf{x}^{\infty})\leq0,\,\forall\mathbf{x}\in\mathcal{B}.
\]
Since $\nabla f_{\epsilon}^{\alpha}(\mathbf{x}^{\infty})=\nabla q_{\alpha}(\mathbf{x}^{\infty}|\mathbf{x}^{\infty})$
by construction, we obtain 
\[
\nabla f_{\epsilon}^{\alpha}(\mathbf{x}^{\infty})^{T}(\mathbf{x}-\mathbf{x}^{\infty})\leq0,\,\forall\mathbf{x}\in\mathcal{B},
\]
implying that $\mathbf{x}^{\infty}$ is a stationary point of the
problem \eqref{eq:convex-constraint}, which is equivalent to the
problem \eqref{eq:SGEP-smooth-Approx} according to Lemma \ref{lem:obj-convex}
and Proposition \ref{prop:max-convex}. 
\end{IEEEproof}
We note that in the above convergence analysis of Algorithm \ref{alg:MM-SGEP},
the leading generalized eigenvector is assumed to be computed exactly
at each iteration. Recall that the Algorithm \ref{alg:steepest_ascent-1}
is not guaranteed to converge to the leading generalized eigenvector
in principle\textcolor{black}{, so if }it\textcolor{black}{{} is applied
to compute the leading generalized eigenvector,} the convergence of
Algorithm \ref{alg:MM-SGEP} to a stationary point is no longer guaranteed.

\section{Numerical Experiments\label{sec:Numerical-Experiments}}

To compare the performance of the proposed algorithms with existing
ones on the sparse generalized eigenvalue problem (SGEP) and some
of its special cases, we present some experimental results in this
section. All experiments were performed on a PC with a 3.20GHz i5-3470
CPU and 8GB RAM.

\subsection{Sparse Generalized Eigenvalue Problem}

In this subsection, we evaluate the proposed IRQM algorithm for the
sparse generalized eigenvalue problem in terms of computational complexity
and the ability to extract sparse generalized eigenvectors. The benchmark
method considered here is the DC-SGEP algorithm proposed in \cite{sriperumbudur2007sparse,sriperumbudur2011majorization},
which is based on D.C. (difference of convex functions) programming
and minorization-maximization (to the best of our knowledge, this
is the only algorithm proposed for this case). The problem that DC-SGEP
solves is just \eqref{eq:SGEP-Approx} with the surrogate function
$g_{p}(x)=\log(1+\left|x\right|/p)$/$\log(1+1/p)$, but the equality
constraint $\mathbf{x}^{T}\mathbf{B}\mathbf{x}=1$ is relaxed to $\mathbf{x}^{T}\mathbf{B}\mathbf{x}\leq1$.
The DC-SGEP algorithm requires solving a convex quadratically constrained
quadratic program (QCQP) at each iteration, which is solved by the
solver Mosek%
\footnote{Mosek, available at http://www.mosek.com/%
} in our experiments. In the experiments, the stopping condition is
$\left|f(\mathbf{x}^{(k+1)})-f(\mathbf{x}^{(k)})\right|/\max\left(1,\left|f(\mathbf{x}^{(k)})\right|\right)\leq10^{-5}$
for both algorithms. For the proposed IRQM algorithm, the smoothing
parameter is set to be $\epsilon=10^{-8}.$

\subsubsection{Computational Complexity }

In this subsection, we compare the computational complexity of the
proposed IRQM Algorithm \ref{alg:MM-SGEP} with the DC-SGEP algorithm.
The surrogate function $g_{p}(x)=\log(1+\left|x\right|/p)$/$\log(1+1/p)$
is used for both algorithms in this experiment. The preconditioned
steepest ascent method given in Algorithm \ref{alg:steepest_ascent-1}
is applied to compute the leading generalized eigenvector at every
iteration of the IRQM algorithm. To illustrate the effectiveness of
the preconditioning scheme employed in Algorithm \ref{alg:steepest_ascent-1},
we also consider computing the leading generalized eigenvector by
invoking Algorithm \ref{alg:steepest_ascent-1} but without preconditioning,
i.e., setting $\mathbf{P}=\mathbf{I}_{n}$. The data matrices $\mathbf{A}\in\mathbf{S}^{n}$
and $\mathbf{B}\in\mathbf{S}_{++}^{n}$ are generated as $\mathbf{A}=\mathbf{C}+\mathbf{C}^{T}$
and $\mathbf{B}=\mathbf{D}^{T}\mathbf{D}$, with \textbf{$\mathbf{C}\in\mathbf{R}^{n\times n}$},\textbf{
$\mathbf{D}\in\mathbf{R}^{1.2n\times n}$} and the entries of both
$\mathbf{C}$ and $\mathbf{D}$ independent, identically distributed
and following $\mathcal{N}(0,1)$. For both algorithms, the initial
point $\mathbf{x}^{(0)}$ is chosen randomly with each entry following
$\mathcal{N}(0,1)$ and then normalized such that $\left(\mathbf{x}^{(0)}\right)^{T}\mathbf{B}\mathbf{x}^{(0)}=1$.
The parameter $p$ of the surrogate function is chosen to be $1$
and the regularization parameter is $\rho=0.1.$ 

\begin{figure}[htbp]
\centering{}\includegraphics[width=0.9\columnwidth]{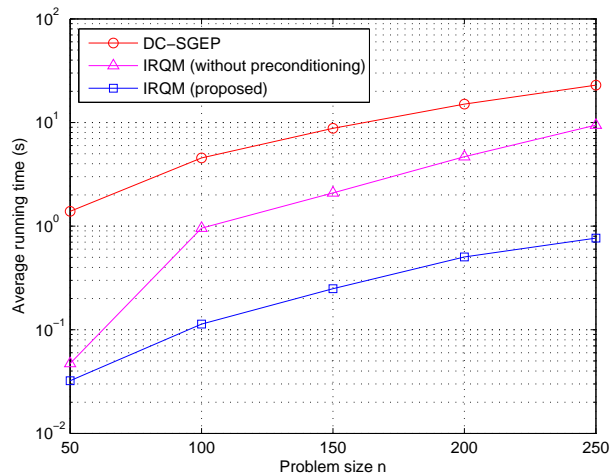}
\protect\caption{\label{fig:cpu_time}Average running time versus problem size. Each
curve is an average of 100 random trials.}
\end{figure}

\begin{figure}[htbp]
\centering{}\includegraphics[width=0.9\columnwidth]{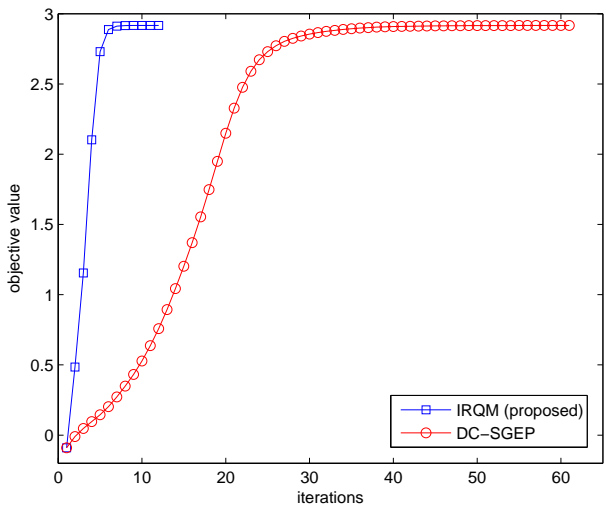}
\protect\caption{\label{fig:objective_evolution}Evolution of the objective function
for one trial with $n=100.$ }
\end{figure}

The computational time for problems with different sizes are shown
in Figure \ref{fig:cpu_time}. The results are averaged over 100 independent
trials. From Figure \ref{fig:cpu_time}, we can see that the preconditioning
scheme is indeed important for the efficiency of Algorithm \ref{alg:steepest_ascent-1}
and the proposed IRQM algorithm is much faster than the DC-SGEP algorithm.
It is worth noting that the solver Mosek which is used to solve the
QCQPs for the DC-SGEP algorithm is well known for its efficiency,
while the IRQM algorithm is entirely implemented in Matlab. The lower
computational complexity of the IRQM algorithm, compared with the
DC-SGEP algorithm, attributes to both the lower per iteration computational
complexity and the faster convergence. To show this, the evolution
of the objective function for one trial with $n=100$ is plotted in
Figure \ref{fig:objective_evolution} and we can see that the proposed
IRQM algorithm takes much fewer iterations to converge. One may also
notice that the two algorithms converge to the same objective value,
but this does not hold in general since the problem is nonconvex.

\subsubsection{Random Data with Underlying Sparse Structure }

In this section, we generate random matrices $\mathbf{A}\in\mathbf{S}^{n}$
and $\mathbf{B}\in\mathbf{S}_{++}^{n}$ such that the matrix pair
$(\mathbf{A},\mathbf{B})$ has a few sparse generalized eigenvectors.
To achieve this, we synthesize the data through the generalized eigenvalue
decomposition $\mathbf{A}=\mathbf{V}^{-T}{\rm Diag}(\mathbf{d})\mathbf{V}^{-1}$
and $\mathbf{B}=\mathbf{V}^{-T}\mathbf{V}^{-1}$, where the first
$k$ columns of $\mathbf{V}\in\mathbf{R}^{n\times n}$ are pre-specified
sparse vectors and the remaining columns are generated randomly, \textbf{$\mathbf{d}$}
is the vector of the generalized eigenvalues. 

Here, we choose $n=100$ and $k=2,$ where the two sparse generalized
eigenvectors are specified as follows 
\[
\begin{cases}
V_{i,1}=\frac{1}{\sqrt{5}} & {\rm for}\quad i=1,\ldots,5,\\
V_{i,1}=0 & {\rm otherwise},
\end{cases}
\]
\vspace{-2bp}
\[
\begin{cases}
V_{i,2}=\frac{1}{\sqrt{5}} & {\rm for}\quad i=6,\ldots,10,\\
V_{i,2}=0 & {\rm otherwise},
\end{cases}
\]
and the generalized eigenvalues are chosen as 
\[
\begin{cases}
d_{1}=10\\
d_{2}=8\\
d_{i}=12, & {\rm for}\quad i=3,4,5\\
d_{i}\sim\mathcal{N}(0,1), & {\rm otherwise}.
\end{cases}
\]

\begin{figure}
\centering{}\includegraphics[width=0.9\columnwidth]{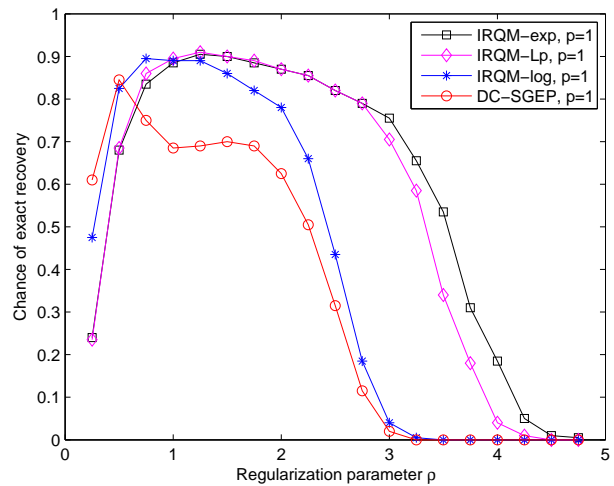}
\protect\caption{\label{fig:recovery_percent_1}Chance of exact recovery versus regularization
parameter $\rho$. Parameter $p=1$ is used for the surrogate functions.}
\end{figure}

\begin{figure}
\centering{}\includegraphics[width=0.9\columnwidth]{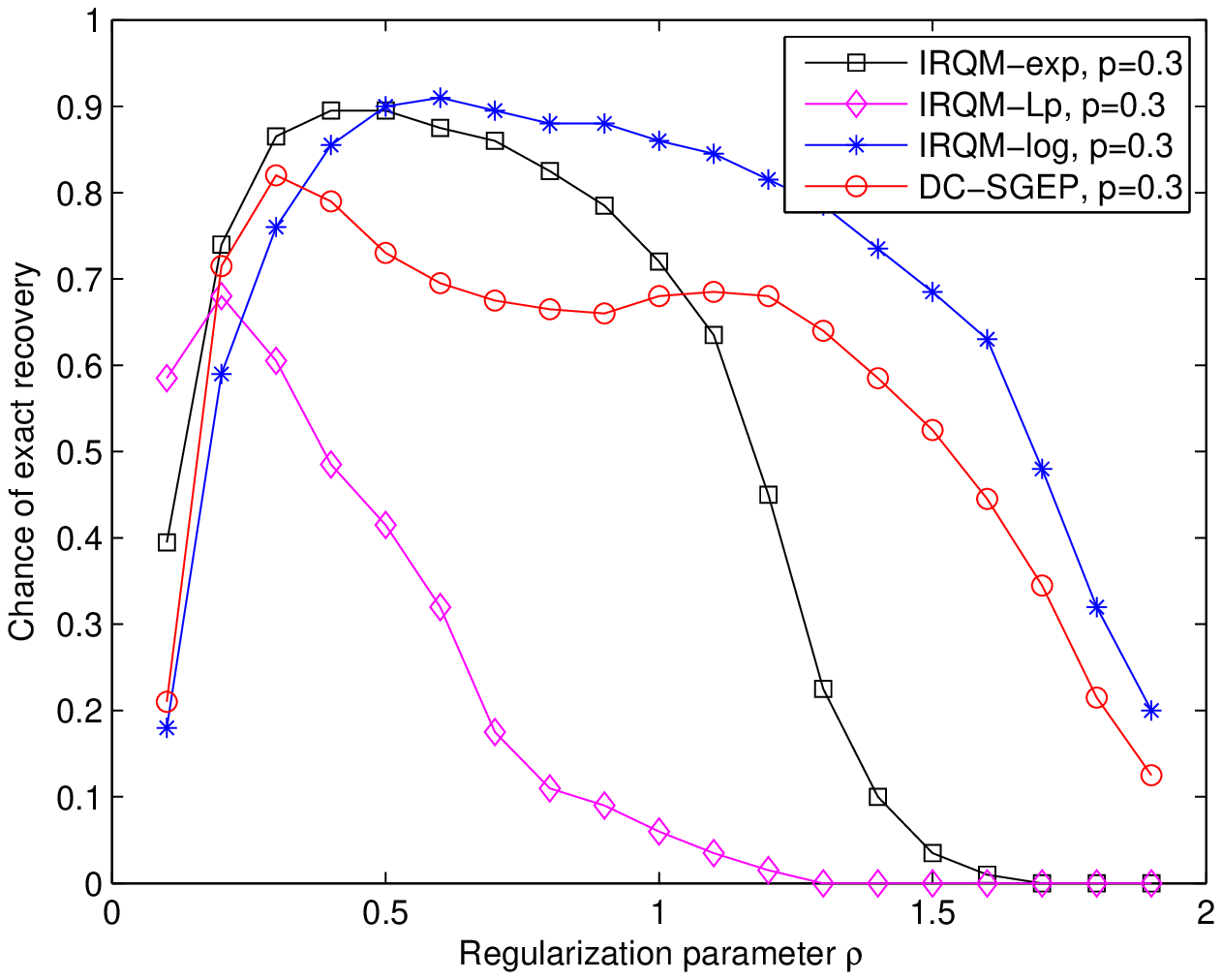}
\protect\caption{\label{fig:recovery_percent_03}Chance of exact recovery versus regularization
parameter $\rho$. Parameter $p=0.3$ is used for the surrogate functions.}
\end{figure}

\begin{figure}
\centering{}\includegraphics[width=0.9\columnwidth]{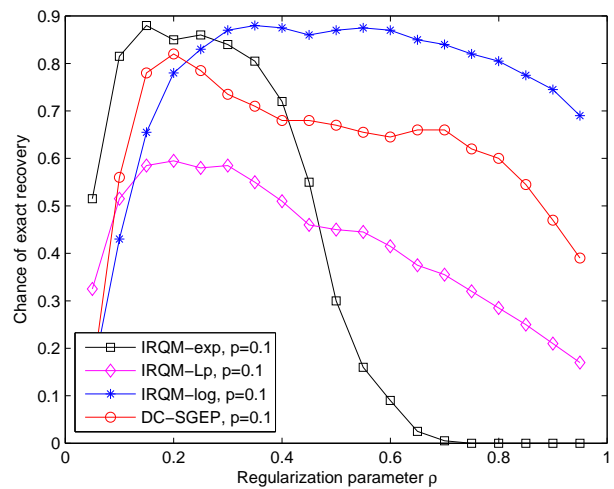}
\protect\caption{\label{fig:recovery_percent_01}Chance of exact recovery versus regularization
parameter $\rho$. Parameter $p=0.1$ is used for the surrogate functions.}
\end{figure}

We generate 200 pairs of $(\mathbf{A},\mathbf{B})$ as described above
and employ the algorithms to compute the leading sparse generalized
eigenvector $\mathbf{x}_{1}\in\mathbf{R}^{100},$ which is hoped to
be close to $\mathbf{V}_{:,1}.$ The underlying sparse generalized
eigenvector $\mathbf{V}_{:,1}$ is considered to be successfully recovered
when $\left\Vert \left|\mathbf{x}_{1}\right|-\mathbf{V}_{:,1}\right\Vert _{2}\leq0.01.$
For the proposed IRQM algorithm, all the three surrogate functions
listed in Table \ref{tab:Smooth-surrogate-functions} are considered
and we call the resulting algorithms ``IRQM-log'', ``IRQM-Lp''
and ``IRQM-exp'', respectively. For all the algorithms, the initial
point $\mathbf{x}^{(0)}$ is chosen randomly. Regarding the parameter
$p$ of the surrogate function, three values, namely $1$, $0.3$
and $0.1$, are compared. The corresponding performance along the
whole path of the regularization parameter $\rho$ is plotted in Fig.
\ref{fig:recovery_percent_1}, \ref{fig:recovery_percent_03} and
\ref{fig:recovery_percent_01}, respectively.

From Fig. \ref{fig:recovery_percent_1}, we can see that for the case
$p=1$, the best chance of exact recovery achieved by the three IRQM
algorithms are very close and all higher than that achieved by the
DC-SGEP algorithm. From Fig. \ref{fig:recovery_percent_03} and \ref{fig:recovery_percent_01},
we can see that as $p$ becomes smaller, the best chance of exact
recovery achieved by IRQM-exp, IRQM-log and DC-SGEP stay almost the
same as in the case $p=1$ (in fact decrease a little bit when $p=0.1$),
but the performance of IRQM-Lp degrades a lot. This may be explained
by the fact that as $p$ becomes smaller, the surrogate function $\left|x\right|^{p}$
tends to the function $\mathrm{sgn}(\left|x\right|)$ much faster
than the other two surrogate functions. So when $p=0.1$ for example,
it is much more pointed and makes the algorithm easily get stuck at
some local point. In this sense, the log-based and exp-based surrogate
functions seem to be better choices as they are not so sensitive to
the choice of $p$.

\subsubsection{Decreasing Scheme of the Smoothing Parameter $\epsilon$}

As have been discussed in the end of Section \ref{sub:Smooth-Approximations},
choosing a relatively large smoothing parameter $\epsilon$ at the
beginning and decreasing it gradually may probably lead to better
performance than the fixed $\epsilon$ scheme. In this section, we
consider such a decreasing scheme and compare its performance with
the fixed $\epsilon$ scheme in which the smoothing parameter is fixed
to be $\epsilon=10^{-8}$. The decreasing scheme that we will adopt
is inspired by the continuation approach in \cite{becker2011NESTA}.
The idea is to apply the IRQM algorithm to a succession of problems
with decreasing smoothing parameters $\epsilon^{(0)}>\epsilon^{(1)}>\cdots>\epsilon^{(T)}$
and solve the intermediate problems with less accuracy, where $T$
is the number of decreasing steps. More specifically, at step $t=0,\ldots,T$,
we apply the IRQM algorithm with smoothing parameter $\epsilon^{(t)}$
and stopping criterion $\left|f(\mathbf{x}^{(k+1)})-f(\mathbf{x}^{(k)})\right|/\max\left(1,\left|f(\mathbf{x}^{(k)})\right|\right)\leq\sqrt{\epsilon^{(t)}}/10$
and then decrease the smoothing parameter for the next step by $\epsilon^{(t+1)}=\gamma\epsilon^{(t)}$
with $\gamma=(\epsilon^{(T)}/\epsilon^{(0)})^{1/T}$. At each step
the IRQM algorithm is initialized with the solution of the previous
step. The initial smoothing parameter is chosen as $\epsilon^{(0)}=\left\Vert \mathbf{x}^{(0)}\right\Vert _{\infty}/4$,
where $\mathbf{x}^{(0)}$ is the random initial point and the minimum
smoothing parameter is set as $\epsilon^{(T)}=10^{-8},$ which is
the parameter used in the fixed $\epsilon$ scheme. The number of
decreasing steps is set to $T=5$ in our experiment.

\begin{figure}[htbp]
\centering{}\includegraphics[width=0.9\columnwidth]{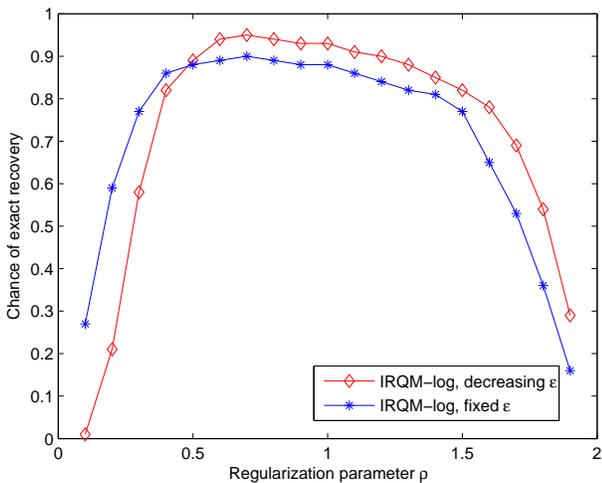}
\protect\caption{\label{fig:recovery_percent_decreasing_epsi}Chance of exact recovery
versus regularization parameter $\rho$. The log-based surrogate function
with parameter $p=0.3$ is used.}
\end{figure}

The remaining settings are the same as in the previous subsection
and the log-based surrogate function with parameter $p=0.3$ is used
for the IRQM algorithm. The performance of the two schemes are shown
in Fig. \ref{fig:recovery_percent_decreasing_epsi}. From the figure,
we can see that the decreasing scheme of the smoothing parameter achieves
a higher chance of exact recovery. 

\subsection{Sparse Principal Component Analysis $(\mathbf{B}=\mathbf{I}_{n})$}

In this section, we consider the special case of the sparse generalized
eigenvalue problem in which the matrix $\mathbf{B}$ is the identity
matrix, i.e., the sparse PCA problem, which has received most of the
recent attention in the literature. In this case, the matrix $\mathbf{A}$
is usually a (scaled) covariance matrix. Although there exists a vast
literature on sparse PCA, most popular algorithms are essentially
variations of the generalized power method (GPower) proposed in \cite{Gpower}.
Thus we choose the GPower methods, namely ${\rm GPower}_{\ell_{1}}$
and ${\rm GPower}_{\ell_{0}}$, as the benchmarks in this section.
The Matlab code of the GPower algorithms was downloaded from the authors'
website. For the proposed Algorithm \ref{alg:approx-penalized-diagonal},
the surrogate function is chosen to be $g_{p}(x)=\left|x\right|,$
such that the penalty function is just the $\ell_{1}$-norm, which
is the same as in ${\rm GPower}_{\ell_{1}}$. We call the resulting
algorithm $"{\rm MM}_{\ell_{1}}"$ and the Algorithm \ref{alg:L0-penalized-diagonal}
is referred to as $"{\rm MM}_{\ell_{0}}"$ in this section. 

Note that for GPower methods, direct access to the original data matrix
$\mathbf{C}$ is required. When only the covariance matrix is available,
a factorization of the form $\mathbf{A}=\mathbf{C}^{T}\mathbf{C}$
is needed (e.g., by eigenvalue decomposition or by Cholesky decomposition).
If the data matrix $\mathbf{C}$ is of size $m\times n$, then the
per-iteration computational cost is $\mathcal{O}(mn)$ for all the
four algorithms under consideration.

\subsubsection{Computational Complexity}

In this subsection, we compare the computational complexity of the
four algorithms mentioned above, i.e., ${\rm GPower}_{\ell_{1}}$,
${\rm GPower}_{\ell_{0}}$, ${\rm MM}_{\ell_{1}}$ and ${\rm MM}_{\ell_{0}}$.
The data matrix $\mathbf{C}\in\mathbf{R}^{n\times n}$ is generated
randomly with the entries independent, identically distributed and
following $\mathcal{N}(0,1)$. The stopping condition is set to be
$\left|f(\mathbf{x}^{(k+1)})-f(\mathbf{x}^{(k)})\right|/\max\left(1,\left|f(\mathbf{x}^{(k)})\right|\right)\leq10^{-5}$
for all the algorithms. The smoothing parameter for algorithm ${\rm MM}_{\ell_{1}}$
is fixed to be $\epsilon=10^{-8}.$ The regularization parameter $\rho$
is chosen such that the solutions of the four algorithms exhibit similar
cardinalities (with about 5\% nonzero entries).

\begin{figure}
\centering{}\includegraphics[width=0.9\columnwidth]{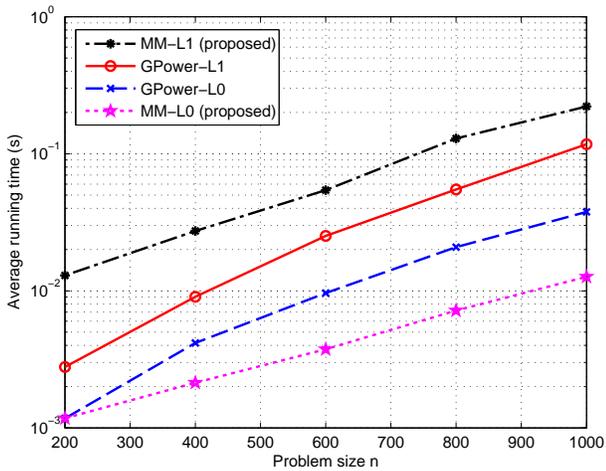}
\protect\caption{\label{fig:cpu_time-SPCA}Average running time versus problem size.
Each curve is an average of 100 random trials.}
\end{figure}

The average running time over 100 independent trials for problems
with different sizes are shown in Figure \ref{fig:cpu_time-SPCA}.
From the figure, we can see that the two $\ell_{0}$-norm penalized
methods are faster than the two $\ell_{1}$-norm penalized methods
and the proposed ${\rm MM}_{\ell_{0}}$ is the fastest among the four
algorithms, especially for problems of large size. For the two $\ell_{1}$-norm
penalized methods, the proposed ${\rm MM}_{\ell_{1}}$ is slower than
${\rm GPower}_{\ell_{1}}$, which may result from the fact that ${\rm MM}_{\ell_{1}}$
minorizes both the quadratic term and the $\ell_{1}$ penalty term
while ${\rm GPower}_{\ell_{1}}$ keeps the $\ell_{1}$ penalty term.
It is worth noting that the ${\rm GPower}_{\ell_{1}}$ is specialized
for $\ell_{1}$ penalty, while Algorithm \ref{alg:approx-penalized-diagonal}
can also deal with various surrogate functions other than the $\ell_{1}$
penalty.

\subsubsection{Random Data Drawn from a Sparse PCA Model}

In this subsection, we follow the procedure in \cite{shen2008sparse}
to generate random data with a covariance matrix having sparse eigenvectors.
To achieve this, we first construct a covariance matrix through the
eigenvalue decomposition $\mathbf{A}=\mathbf{V}{\rm Diag}(\mathbf{d})\mathbf{V}^{T}$,
where the first $k$ columns of $\mathbf{V}\in\mathbf{R}^{n\times n}$
are pre-specified sparse orthonormal vectors. A data matrix $\mathbf{C}\in\mathbf{R}^{m\times n}$
is then generated by drawing $m$ samples from a zero-mean normal
distribution with covariance matrix $\mathbf{A}$, that is, $\mathbf{C}\sim\mathcal{N}(\mathbf{0},\mathbf{A}).$

Following the settings in \cite{Gpower}, we choose $n=500,$ $k=2,$
and $m=50$, where the two orthonormal eigenvectors are specified
as follows 
\[
\begin{cases}
V_{i,1}=\frac{1}{\sqrt{10}} & {\rm for}\quad i=1,\ldots,10,\\
V_{i,1}=0 & {\rm otherwise},
\end{cases}
\]
\vspace{-2bp}
\[
\begin{cases}
V_{i,2}=\frac{1}{\sqrt{10}} & {\rm for}\quad i=11,\ldots,20,\\
V_{i,2}=0 & {\rm otherwise}.
\end{cases}
\]
The eigenvalues are fixed at $d_{1}=400,$ $d_{2}=300$ and $d_{i}=1,$
for $i=3,\ldots,500$.

\begin{figure}[htbp]
\centering{}\includegraphics[width=0.88\columnwidth]{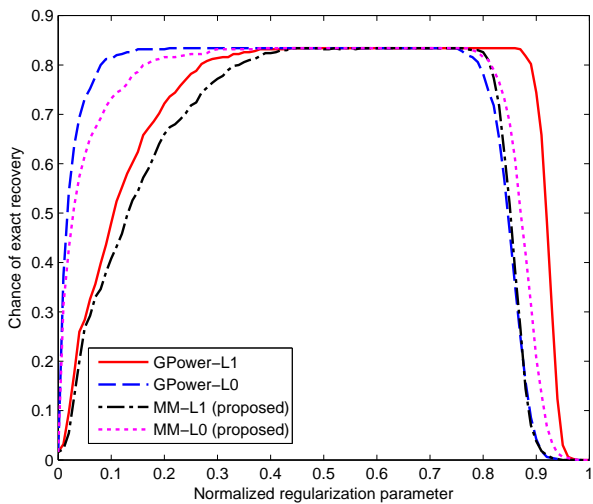}
\protect\caption{\label{fig:SPCA_recovery}Chance of exact recovery versus normalized
regularization parameter.}
\end{figure}

We randomly generate 500 data matrices $\mathbf{C}\in\mathbf{R}^{m\times n}$
and employ the four algorithms to compute the leading sparse eigenvector
$\mathbf{x}_{1}\in\mathbf{R}^{500},$ which is hoped to recover $\mathbf{V}_{:,1}.$
We consider the underlying sparse eigenvector $\mathbf{V}_{:,1}$
is successfully recovered when $\left|\mathbf{x}_{1}^{T}\mathbf{V}_{:,1}\right|>0.99.$
The chance of successful recovery over a wide range of regularization
parameter $\rho$ is plotted in Figure \ref{fig:SPCA_recovery}. The
horizontal axis shows the normalized regularization parameter, that
is $\rho/\max_{i}\left\Vert \mathbf{C}_{:,i}\right\Vert _{2}$ for
${\rm GPower}_{\ell_{1}}$ and $\rho/\max_{i}\left\Vert \mathbf{C}_{:,i}\right\Vert _{2}^{2}$
for ${\rm GPower}_{\ell_{0}}$and ${\rm MM}_{\ell_{0}}.$ For ${\rm MM}_{\ell_{1}}$
algorithm, we use $\rho/\left(2\left\Vert \mathbf{C}^{T}\mathbf{C}\right\Vert _{\infty,2}\right),$
where $\left\Vert \cdot\right\Vert _{\infty,2}$ is the operator norm
induced by $\left\Vert \cdot\right\Vert _{\infty}$ and $\left\Vert \cdot\right\Vert _{2}$.
From the figure, we can see that the highest chance of exact recovery
achieved by the four algorithms is the same and for all algorithms
it is achieved over a relatively wide range of $\rho$.

\subsubsection{Gene Expression Data }

DNA microarrays allow measuring the expression level of thousands
of genes at the same time and this opens the possibility to answer
some complex biological questions. But the amount of data created
in an experiment is usually large and this makes the interpretation
of these data challenging. PCA has been applied as a tool in the studies
of gene expression data and their interpretation \cite{riva2005DNA}.
Naturally, sparse PCA, which extracts principal components with only
a few nonzero elements can potentially enhance the interpretation. 

\begin{figure}[htbp]
\centering{}\includegraphics[width=0.9\columnwidth]{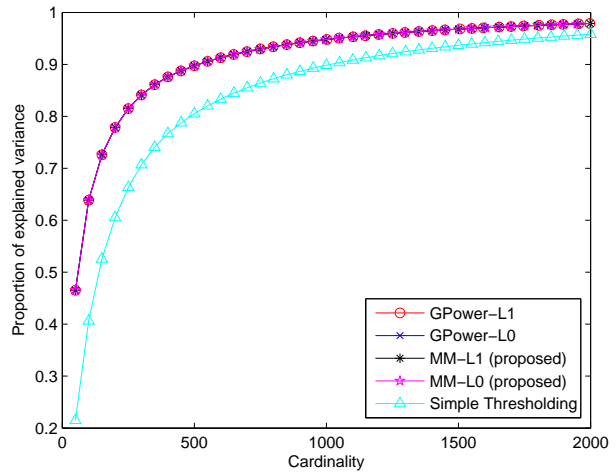}
\protect\caption{\label{fig:Gene-expression}Trade-off curves between explained variance
and cardinality. }
\end{figure}

In this subsection, we test the performance of the algorithms on gene
expression data collected in the breast cancer study by Bild et al.
\cite{bild2006Gene}. The data set contains 158 samples over 12625
genes, resulting in a $158\times12625$ data matrix. Figure \ref{fig:Gene-expression}
shows the explained variance versus cardinality for five algorithms,
including the simple thresholding scheme. The proportion of explained
variance is computed as $\mathbf{x}_{{\rm SPCA}}^{T}(\mathbf{C}^{T}\mathbf{C})\mathbf{x}_{{\rm SPCA}}/\mathbf{x}_{{\rm PCA}}^{T}(\mathbf{C}^{T}\mathbf{C})\mathbf{x}_{{\rm PCA}},$
where $\mathbf{x}_{{\rm SPCA}}$ is the sparse eigenvector extracted
by sparse PCA algorithms, $\mathbf{x}_{{\rm PCA}}$ is the true leading
eigenvector and $\mathbf{C}$ is the data matrix. The simple thresholding
scheme first computes the regular principal component $\mathbf{x}_{{\rm PCA}}$
and then keeps a required number of entries with largest absolute
values. From the figure, we can see that the proportion of variance
being explained increases as the cardinality increases as expected.
For a fixed cardinality, the two GPower algorithms and the two proposed
MM algorithms can explain almost the same amount of variance, all
higher than the simple thresholding scheme, especially when the cardinality
is small. 

\section{Conclusions\label{sec:Conclusions}}

We have developed an efficient algorithm IRQM that allows to obtain
sparse generalized eigenvectors of a matrix pair. After approximating
the $\ell_{0}$-norm penalty by some nonconvex surrogate functions,
the minorization-maximization scheme is applied and the sparse generalized
eigenvalue problem is turned into a sequence of regular generalized
eigenvalue problems. The convergence to a stationary point is proved.
Numerical experiments show that the proposed IRQM algorithm outperforms
an existing algorithm based on D.C. programming in terms of both computational
cost and support recovery. For sparse generalized eigenvalue problems
with special structure (but still including sparse PCA as a special
instance), two more efficient algorithms that have a closed-form solution
at every iteration are derived again based on the minorization-maximization
scheme. On both synthetic random data and real-life gene expression
data, the two algorithms are shown experimentally to have similar
performance to the state-of-the-art.

\appendices{}

\section{Proof of Proposition \ref{prop:suboptimal} \label{sec:Proof-of-Proposition_suboptimal}}
\begin{IEEEproof}
From Lemma \ref{lem:Low-up_bound}, it is easy to show that 
\begin{equation}
f_{\epsilon}(\mathbf{x})\geq f(\mathbf{x})\geq f_{\epsilon}(\mathbf{x})-\rho n\left(g_{p}(\epsilon)-\frac{g_{p}^{\prime}(\epsilon)}{2}\epsilon\right),\,\forall\mathbf{x}\in\mathbf{R}^{n}.\label{eq:bound1}
\end{equation}
Since problems \eqref{eq:SGEP-Approx} and \eqref{eq:SGEP-smooth-Approx}
have the same constraint set, we have 
\begin{equation}
f_{\epsilon}(\mathbf{x}_{\epsilon}^{\star})\geq f(\mathbf{x}^{\star}).\label{eq:bound2}
\end{equation}
From the fact that $\mathbf{x}^{\star}$ is a global maximizer of
problem \eqref{eq:SGEP-Approx}, we know 
\begin{equation}
f(\mathbf{x}^{\star})\geq f(\mathbf{x}_{\epsilon}^{\star}).\label{eq:bound3}
\end{equation}
Combining \eqref{eq:bound1}, \eqref{eq:bound2} and \eqref{eq:bound3},
yields
\[
f_{\epsilon}(\mathbf{x}_{\epsilon}^{\star})\geq f(\mathbf{x}^{\star})\geq f(\mathbf{x}_{\epsilon}^{\star})\geq f_{\epsilon}(\mathbf{x}_{\epsilon}^{\star})-\rho n\left(g_{p}(\epsilon)-\frac{g_{p}^{\prime}(\epsilon)}{2}\epsilon\right).
\]
Thus, 
\[
0\leq f(\mathbf{x}^{\star})-f(\mathbf{x}_{\epsilon}^{\star})\leq\rho n\left(g_{p}(\epsilon)-\frac{g_{p}^{\prime}(\epsilon)}{2}\epsilon\right).
\]

Since $g_{p}(\cdot)$ is concave and monotone increasing on $[0,+\infty)$,
it is easy to show that $g_{p}(\epsilon)\geq g_{p}^{\prime}(\epsilon)\epsilon\geq0$,
for any $\epsilon>0.$ Hence 
\begin{equation}
g_{p}(\epsilon)\geq g_{p}(\epsilon)-\frac{g_{p}^{\prime}(\epsilon)}{2}\epsilon\geq0.\label{eq:bounded}
\end{equation}
Since $g_{p}(\cdot)$ is continuous and monotone increasing on $[0,+\infty)$
and $g_{p}(0)=0,$ we have $\lim_{\epsilon\downarrow0}g_{p}(\epsilon)=0.$
Together with \eqref{eq:bounded}, we can conclude that 
\[
\lim_{\epsilon\downarrow0}\left(g_{p}(\epsilon)-\frac{g_{p}^{\prime}(\epsilon)}{2}\epsilon\right)=0.
\]
Since $\rho$ and $n$ are constants, the proof is complete.
\end{IEEEproof}

\section{Proof of Proposition \ref{prop:nonconvex-QCQP} \label{sec:Proof-of-Proposition_QCQP}}
\begin{IEEEproof}
First notice that the problem \eqref{eq:non-convex-QCQP-2} is a nonconvex
QCQP but with only one constraint, thus the strong duality holds \cite{huang2007complex,huang2010rank}.
The optimality conditions for this problem are 
\begin{eqnarray}
\left(\mu{\rm Diag}(\mathbf{b})+\rho{\rm Diag}(\mathbf{w})\right)\mathbf{x} & = & \mathbf{a}\label{eq:optimal-1-1}\\
\mathbf{x}^{T}{\rm Diag}(\mathbf{b})\mathbf{x} & = & 1\label{eq:optimal-2-1}\\
\mu{\rm Diag}(\mathbf{b})+\rho{\rm Diag}(\mathbf{w}) & \succeq & \mathbf{0}.\label{eq:optimal-3-1}
\end{eqnarray}
Let us define 
\begin{equation}
\mathcal{I}_{{\rm min}}=\arg\min\{\rho w_{i}/b_{i}:i\in\{1,\ldots,n\}\}
\end{equation}
and 
\begin{equation}
\mu_{{\rm min}}=-\min\{\rho w_{i}/b_{i}:i\in\{1,\ldots,n\}\}.
\end{equation}
Then the third optimality condition \eqref{eq:optimal-3-1} is just
$\mu\geq\mu_{{\rm min}},$ since $w_{i}>0,$ $\rho>0,$ $b_{i}>0$.
Let us consider the optimality condition in two different cases:

1) $\mu>\mu_{{\rm min}}$. In this case, $\mu{\rm Diag}(\mathbf{b})+\rho{\rm Diag}(\mathbf{w})\succ\mathbf{0}$,
from the first optimality condition \eqref{eq:optimal-1-1} we get
\begin{equation}
\mathbf{x}=\left({\rm Diag}(\mu\mathbf{b}+\rho\mathbf{w})\right)^{-1}\mathbf{a}.\label{eq:x-equation-2}
\end{equation}
Substituting it into the second optimality condition \eqref{eq:optimal-2-1},
yields  
\begin{equation}
\sum_{i\in\mathcal{I}_{{\rm min}}}\frac{b_{i}a_{i}^{2}}{(\mu b_{i}+\rho w_{i})^{2}}+\sum_{i\notin\mathcal{I}_{{\rm min}}}\frac{b_{i}a_{i}^{2}}{(\mu b_{i}+\rho w_{i})^{2}}=1,\label{eq:mu-equation-1-1}
\end{equation}
and it is easy to see that the left hand side is monotonically decreasing
for $\mu\in(\mu_{{\rm min}},+\infty)$. If $\exists i\in\mathcal{I}_{{\rm min}}$,
such that $a_{i}^{2}>0$, then the left hand side of \eqref{eq:mu-equation-1-1}
tends to $+\infty$ as $\mu\rightarrow\mu_{{\rm min}}.$ Notice that
the left hand side goes to $0$ as $\mu\rightarrow+\infty,$ thus
we are guaranteed to find a $\mu\in(\mu_{{\rm min}},+\infty)$ satisfying
equation \eqref{eq:mu-equation-1-1}. In practice, we may use bisection
method to find the value of $\mu$. If $a_{i}^{2}=0$, $\forall i\in\mathcal{I}_{{\rm min}}$,
there still exists a $\mu>\mu_{{\rm min}}$ that satisfies equation
\eqref{eq:mu-equation-1-1} if and only if 
\begin{equation}
\sum_{i\notin\mathcal{I}_{{\rm min}}}\frac{b_{i}a_{i}^{2}}{(\mu_{{\rm min}}b_{i}+\rho w_{i})^{2}}>1.\label{eq:condition1-1}
\end{equation}
If \eqref{eq:condition1-1} does not hold, it implies $\mu=\mu_{{\rm min}}.$ 

2) $\mu=\mu_{{\rm min}}.$ In this case, we cannot compute $\mathbf{x}$
via equation \eqref{eq:x-equation-2} anymore. Then to obtain $\mathbf{x}$,
we first notice from \eqref{eq:optimal-1-1} that 
\begin{equation}
x_{i}=\frac{a_{i}}{\mu_{{\rm min}}b_{i}+\rho w_{i}},\,\,\forall i\notin\mathcal{I}_{{\rm min}}.
\end{equation}
Then, according to equation \eqref{eq:optimal-2-1}, for $x_{i}$,
$i\in\mathcal{I}_{{\rm min}}$, they just need to satisfy the following
equation
\begin{equation}
\sum_{i\in\mathcal{I}_{{\rm min}}}b_{i}x_{i}^{2}=1-\sum_{i\notin\mathcal{I}_{{\rm min}}}\frac{b_{i}a_{i}^{2}}{(\mu_{{\rm min}}b_{i}+\rho w_{i})^{2}}.\label{eq:x-equation-1-1}
\end{equation}
When ${\rm card}(\mathcal{I}_{{\rm min}})>1,$ \eqref{eq:x-equation-1-1}
has infinite number of solutions and we may choose arbitrary one.
\end{IEEEproof}

\section{Proof of Proposition \ref{prop:L0-closed-form} \label{sec:Proof-of-Proposition_L0_closed_form}}
\begin{IEEEproof}
The problem \eqref{eq:L0-penalized} can be rewritten as 
\[
\underset{s\in\{1,\ldots,n\}}{\mathsf{maximize}}\left\{ -\rho s+\underset{\mathbf{x}}{\mathsf{max}}\left\{ \mathbf{a}^{T}\mathbf{x}:\left\Vert \mathbf{x}\right\Vert _{2}=1,\left\Vert \mathbf{x}\right\Vert _{0}\leq s\right\} \right\} .
\]
The inner maximization has a closed-form solution 
\[
x_{i}^{\star}=\begin{cases}
a_{i}/\sqrt{\sum_{j=1}^{s}a_{j}^{2}}, & i\leq s\\
0 & {\rm otherwise},
\end{cases}
\]
then the problem becomes 
\[
\underset{s\in\{1,\ldots,n\}}{\mathsf{maximize}}\left\{ -\rho s+\sqrt{{\textstyle \sum_{i=1}^{s}}a_{i}^{2}}\right\} .
\]
It's easy to see that the optimal $s$ is the largest integer $p$
that satisfies the following inequality
\begin{equation}
\sqrt{{\textstyle \sum_{i=1}^{p}}a_{i}^{2}}>\sqrt{{\textstyle \sum_{i=1}^{p-1}}a_{i}^{2}}+\rho.\label{eq:L0-inequal-1}
\end{equation}
By squaring both sides of this inequality, we get 
\[
a_{p}^{2}>\rho^{2}+2\rho\sqrt{{\textstyle \sum_{i=1}^{p-1}}a_{i}^{2}},
\]
which means $\left|a_{p}\right|>\rho$ is a necessary condition for
\eqref{eq:L0-inequal-1} to be satisfied. Thus, in practice to find
the largest integer $p$ that satisfies \eqref{eq:L0-inequal-1} we
only need to check for all $p$'s with $\left|a_{p}\right|>\rho$.
If $0<\left|a_{1}\right|\leq\rho$, it is easy to see that the solution
of the problem \eqref{eq:L0-penalized} is given by \eqref{eq:a_less_rho}. 
\end{IEEEproof}

\section{Proof of Lemma \ref{lem:Lipschitz_gradient} \label{sub:Proof-of-Lemma_Lipschitz}}
\begin{IEEEproof}
From the way $g_{p}^{\epsilon}(x)$ is constructed, it is continuously
differentiable. It remains to show that the gradient 
\[
\left(g_{p}^{\epsilon}\right)^{\prime}(x)=\begin{cases}
\frac{g_{p}^{\prime}(\epsilon)}{\epsilon}x, & \left|x\right|\leq\epsilon\\
g_{p}^{\prime}(x), & \left|x\right|>\epsilon
\end{cases}
\]
 is Lipschitz continuous. From the fact that $g_{p}(x)$ is concave
and monotone increasing on $(0,+\infty)$, we know that $g_{p}^{\prime}(x)$
is non-increasing on $(0,+\infty)$ and $g_{p}^{\prime}(x)>0.$ Since
$g_{p}^{\epsilon}(x)$ is an even function, $\left(g_{p}^{\epsilon}\right)^{\prime}(x)$
is odd. Thus, $\left(g_{p}^{\epsilon}\right)^{\prime}(x)$ is non-increasing
on $(-\infty,-\epsilon),$ linearly increasing on $[-\epsilon,\epsilon]$
and non-increasing on $(\epsilon,+\infty).$ In addition, $\left(g_{p}^{\epsilon}\right)^{\prime}(x)\leq0$
when $x<0$ and $\left(g_{p}^{\epsilon}\right)^{\prime}(x)\geq0$
when $x>0.$ With $\left(g_{p}^{\epsilon}\right)^{\prime}(x)$ having
these properties, to show the Lipschitz continuity of $\left(g_{p}^{\epsilon}\right)^{\prime}(x)$,
it is sufficient to show that $\left(g_{p}^{\epsilon}\right)^{\prime}(x)$
is Lipschitz continuous on $[-\epsilon,\epsilon]$ and $(\epsilon,+\infty)$
respectively. 

On $[-\epsilon,\epsilon]$, $\left(g_{p}^{\epsilon}\right)^{\prime}(x)=\frac{g_{p}^{\prime}(\epsilon)}{\epsilon}x$,
which is Lipschitz continuous with Lipschitz constant $\frac{g_{p}^{\prime}(\epsilon)}{\epsilon}.$ 

On $(\epsilon,+\infty)$, $\left(g_{p}^{\epsilon}\right)^{\prime}(x)=g_{p}^{\prime}(x)$,
from Assumption \ref{assump} we know that $g_{p}^{\prime}(x)$ is
convex and differentiable on $(0,+\infty)$. Since $g_{p}^{\prime}(x)$
is also non-increasing, we can conclude that $g_{p}^{\prime\prime}(x)\leq0$
and is non-decreasing on $(0,+\infty).$ Thus, on $(\epsilon,+\infty)$,
$\left|g_{p}^{\prime\prime}(x)\right|$ is bounded by $\left|g_{p}^{\prime\prime}(\epsilon)\right|$
and the Lipschitz continuity of $\left(g_{p}^{\epsilon}\right)^{\prime}(x)$
on $(\epsilon,+\infty)$ follows. 
\end{IEEEproof}
\bibliographystyle{IEEEtran}
\bibliography{SPCA}

\end{document}